\documentclass[journal]{IEEEtran}
\usepackage{cite}

\ifCLASSINFOpdf
\else
\fi
\usepackage{graphicx}

\usepackage{amsmath,amssymb}
\usepackage{algorithmic,algorithm}

\usepackage{array}
\usepackage{url}

\usepackage{color}
\usepackage{xcolor}
\usepackage{booktabs}
\usepackage{wrapfig}
\usepackage{tabularx}

\usepackage{amsmath,amssymb}
\usepackage{multirow}
\usepackage{wrapfig}
\usepackage{color}
\usepackage{colortbl}
\usepackage{tabularx}
\usepackage{url}
\usepackage{float}
\usepackage{booktabs}
\usepackage{hyperref}
\usepackage{caption} 
\usepackage{subcaption}
\usepackage{bbding}
\usepackage{xcolor}
\usepackage{array}
\usepackage{makecell}
\usepackage{ulem}

\definecolor{Salmon}{RGB}{250, 128, 114}

\begin{document}
%
\title{ViLAaD: Enhancing ``Attracting and Dispersing'' Source-Free Domain Adaptation \\ with Vision-and-Language Model}
%
%
%

\author{Shuhei~Tarashima~~~~
        Xinqi~Shu~~~~
        Norio~Tagawa

\thanks{
S. Tarashima is with NTT Communications Corporation, Tokyo, Japan.
S. Tarashima, X. Shu and N. Tagawa are with Tokyo Metropolitan University, Tokyo, Japan. 
(E-mail: tarashima@acm.org)
}
\thanks{
This work has been submitted to the IEEE for possible publication. Copyright may be transferred without notice, after which this version may no longer be accessible.
}

}

\maketitle

\begin{abstract}
Source-Free Domain Adaptation (SFDA) aims to adapt a pre-trained source model to a target dataset from a different domain without access to the source data. 
Conventional SFDA methods are limited by the information encoded in the pre-trained source model and the unlabeled target data. 
Recently, approaches leveraging auxiliary resources have emerged, yet remain in their early stages, offering ample opportunities for research.
In this work, we propose a novel method that incorporates auxiliary information by extending an existing SFDA framework using Vision-and-Language (ViL) models. 
Specifically, we build upon Attracting and Dispersing (AaD), a widely adopted SFDA technique, and generalize its core principle to naturally integrate ViL models as a powerful initialization for target adaptation.
Our approach, called ViL-enhanced AaD (ViLAaD), preserves the simplicity and flexibility of the AaD framework, while leveraging ViL models to significantly boost adaptation performance.
We validate our method through experiments using various ViL models, demonstrating that ViLAaD consistently outperforms both AaD and zero-shot classification by ViL models, especially when both the source model and ViL model provide strong initializations.
Moreover, the flexibility of ViLAaD allows it to be seamlessly incorporated into an alternating optimization framework with ViL prompt tuning and extended with additional objectives for target model adaptation. 
Extensive experiments on four SFDA benchmarks show that this enhanced version, ViLAaD++, achieves state-of-the-art performance across multiple SFDA scenarios, including Closed-set SFDA, Partial-set SFDA, and Open-set SFDA.
\end{abstract}


%
\IEEEpeerreviewmaketitle

\section{Introduction}
\label{sec:intro}
\IEEEPARstart{M}{odels} trained under a supervised learning paradigm perform well when the target domain closely matches the source domain used for training. 
However, their performance significantly degrades under {\it domain shift} \cite{quionero-candela+2009} between source and target data. 
Domain Adaptation (DA) addresses this issue by reducing the discrepancy between a labeled source domain and an unlabeled target domain.
Among various DA settings, Source-Free Domain Adaptation (SFDA) \cite{liang+2020icml} has gained significant attention. 
In SFDA, the source model is adapted to the target dataset \textit{without access to the source data}, making it more practical in real-world scenarios where source data may be unavailable due to privacy concerns or resource limitations.
\begin{figure}[t]
\centering
\includegraphics[width=0.48\textwidth, page=2]{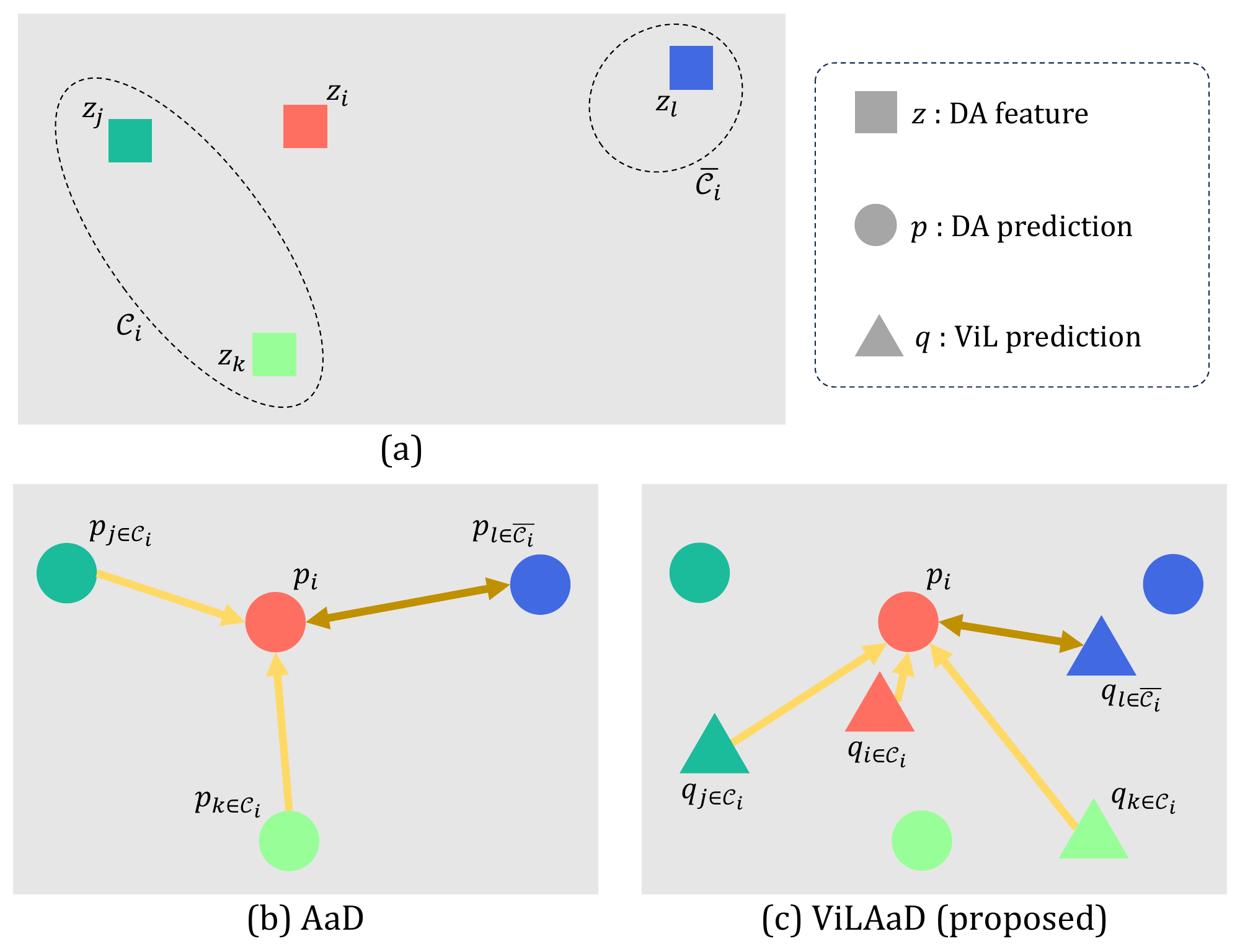}
\caption{
Suppose we have a domain adaptation (DA) model and a target dataset. 
As illustrated in (a), for the $i$-th target example, the $j$-th and $k$-th examples are part of its closed neighbor set $\mathcal{C}_{i}$ in the feature space of the DA model, while the $l$-th example belongs to its complementary set $\bar{\mathcal{C}}_{i}$. 
AaD \cite{yang+2022neurips} encourages the DA model to produce similar predictions for the $i$-th example and its neighbors in $\mathcal{C}_{i}$, while pushing apart the predictions for the $i$-th example and those in $\bar{\mathcal{C}}_{i}$ (as shown in (b)). 
In contrast, our proposed ViLAaD method enhances the alignment between the DA model’s prediction for the $i$-th example and the Vision-and-Language (ViL) model’s predictions for its neighbors in $\mathcal{C}_{i}$, while still enforcing dissimilarity with the examples in $\bar{\mathcal{C}}_{i}$ (see (c)).
}
\label{fig:diff}
\end{figure}
\par
Since its introduction in 2020 \cite{liang+2020icml}, numerous SFDA methods have been proposed \cite{li+2024tpami,fang+2024nn}. 
Despite continuous advancements, conventional SFDA approaches remain constrained by the knowledge embedded in the pre-trained source model and the unlabeled target data. 
To overcome these limitations, recent works have explored \textit{auxiliary resources} such as minimal human labeling \cite{lyu+2024eccv,song+2024eccv}, less-biased feature extractors \cite{zhang+2023iccv,tejero+2024eccv}, and Vision-and-Language (ViL) models \cite{zhang+2024ijcv,tang+2024cvpr,tang+2025iclr}. 
While these approaches have shown promising results, the exploration of auxiliary resources in SFDA remains in its early stages, presenting ample research opportunities.
\par
Building on these observations, we introduce a novel approach to leverage auxiliary information by extending an existing SFDA method with ViL models.
Among various SFDA techniques, we select Attracting and Dispersing (AaD) \cite{yang+2022neurips}, a widely adopted method known for its simplicity and strong performance \cite{zhang+2023iccv,mitsuzumi+2024cvpr,xia+2024cvpr,diamant+2024eccv,zhang+2024ijcv}.
As illustrated in Figure \ref{fig:diff} (a) and (b), AaD operates on the principle that samples close in the target model's feature space should produce more similar predictions than those farther apart. 
In this work, we extend AaD by incorporating a ViL model, which serves as a strong prior for the target dataset. 
Specifically, we generalize the core principle of AaD as follows:
{\it
If both the source model and the ViL model provide reasonable initializations for the target dataset, then the target model’s predictions for a given sample should be more aligned with the ViL predictions of nearby samples (including itself) in the target model’s feature space, rather than with the ViL predictions of more distant samples.
}
This concept is illustrated in Figure \ref{fig:diff} (c).
\par
Based on this insight, we propose a novel SFDA algorithm, ViL-enhanced AaD (ViLAaD), which naturally extends AaD in the presence of a ViL model. 
ViLAaD retains the simplicity and flexibility of AaD while leveraging the ViL model to enhance adaptation performance.
We validate our concept by conducting experiments using five ViL models, including CLIP \cite{radford+2021arxiv}, ALBEF \cite{li+2021neurips}, and BLIP \cite{li+2022icml}. 
Results demonstrate that ViLAaD outperforms AaD and zero-shot classification by ViL models when both the source and ViL models serve as effective initializations for the target data (see \S \ref{sec:eval:vilaad}). 
Furthermore, ViLAaD’s flexibility allows seamless integration into an alternating optimization framework with ViL prompt tuning \cite{tang+2024cvpr}, and enables target model adaptation with additional objectives. 
We evaluate this enhanced version, ViLAaD++, through extensive experiments on four SFDA benchmarks: Office-31 \cite{saenko+eccv2010}, Office-Home \cite{venkateswara+2017cvpr}, VisDA-C \cite{peng+2017arxiv,peng+2018cvprw}, and DomainNet126 \cite{peng+2019iccv}.
Experimental results show that ViLAaD++ achieves state-of-the-art performance in many cases of multiple SFDA scenarios, namely Closed-set SFDA, Partial-set SFDA, and Open-set SFDA (see \S \ref{sec:eval:vilaadpp} and \ref{sec:eval:po}).
Our contributions can be summarized as follows:
\begin{itemize}
\item We propose ViL-enhanced AaD (ViLAaD), an extension of AaD \cite{yang+2022neurips} that leverages ViL models for improved adaptation. 
Experimental results demonstrate that ViLAaD surpasses AaD and zero-shot classification by ViL models when both the source and ViL models provide effective initializations for the target data.
\item We introduce ViLAaD++, which builds upon ViLAaD by incorporating:
(1) an alternating optimization framework with ViL prompt tuning, and (2) additional adaptation objectives for refining the target model. Extensive experiments on SFDA benchmark datasets confirm that ViLAaD++ achieves state-of-the-art performance across multiple SFDA scenarios.
\end{itemize}
\section{Related Work}
\label{sec:related}
\subsection{Source-Free Domain Adaptation (SFDA)}
\label{sec:related:sfda}
SFDA has been explored across various Computer Vision (CV) and Machine Learning (ML) tasks, including object detection \cite{li+2021aaai,vs+2023cvpr,lin+2023icme,liu+2023iccv,chen+2023acmmm}, semantic segmentation \cite{rizzoli+2025wacv,li+2025aaai}, multi-object tracking \cite{segu+2023iccv,mancusi+2024neurips}, gesture recognition \cite{guo+2024neurips} and video understanding \cite{zara+2023iccv}.
In this work, we focus on SFDA for image classification \cite{liang+2020icml,mori+2025wacv}, one of the most fundamental and widely extendable tasks in CV and ML. 
Over the years, various SFDA approaches for image classification have been proposed, including
clustering-based methods \cite{li+2021acmmm,yang+2021iccv,yang+2021neurips,tang+2022nn,yang+2022neurips,zhang+2022neurips,wan+2024cvpr}, 
distillation-based methods \cite{tang+2021iros,zhang+2023iccv,xia+2024cvpr,xing+2024eccv}, 
techniques leveraging data augmentation \cite{hwang+2024iclr,mitsuzumi+2024cvpr} or generative models \cite{xia+2021iccv,khramtsova+2024eccv}, 
and methods generating pseudo source domains \cite{huang+2022neurips,tian+2022tcsvt,du+2022aaai,ding+2022cvpr}, pseudo-labels \cite{kim+2021tai,liang+2020icml,liang+2021tpami,lee+2022icml,litrico+2023cvpr,rai+2025wacv,schlachter+2025wacv} or prototypes for target classes \cite{qiu+2021ijcai}. 
For a more comprehensive overview, we refer readers to recent surveys on SFDA \cite{li+2024tpami,fang+2024nn}. 
However, as mentioned in \S\ref{sec:intro}, conventional SFDA methods remain constrained by their reliance on the source model and target data, limiting their adaptability in more complex scenarios.
\par
Recent studies have proposed the use of {\it auxiliary resources} to address this limitation. 
For instance, RobustNN \cite{tejero+2024eccv} incorporates a less-biased feature extractor ({\it e.g.}, an ImageNet-1K classifier) as a ``second-opinion'' to enhance pseudo-labeling under class distribution shifts between source and target domains.
Co-learn \cite{zhang+2023iccv} and DCPL \cite{diamant+2024eccv} extract image features from a pre-trained feature extractor and input them into a nearest-centroid classifier, weighted by adaptation model predictions, to improve pseudo-labeling. 
In contrast, LFTL \cite{lyu+2024eccv} introduces minimal manual annotations for target samples, which are actively selected by the model from previous training epochs to refine adaptation.
RLD \cite{song+2024eccv} aims to balance supervised signals by incorporating latent defending samples during adaptation, thereby reducing the impact of biased user feedback. 
Additionally, POUF \cite{tanwisuth+2023icml} and ReCLIP \cite{xuefeng+2024wacv} directly adapt the well-known Vision-and-Language (ViL) model, CLIP \cite{radford+2021arxiv}, to the target domain.
Co-learn++ \cite{zhang+2024ijcv}, an extension of Co-learn \cite{zhang+2023iccv}, utilizes multiple text prompts with CLIP's text encoder to further enhance pseudo-labeling quality.
DIFO \cite{tang+2024cvpr} and ProDe \cite{tang+2025iclr} leverage a frozen CLIP model to generate pseudo-labels for target data and employ an alternating optimization framework to refine both the learnable prompt in CLIP’s text encoder and the domain-adapted model.
\par
In this work, we leverage ViL models to enhance a prominent SFDA approach, AaD \cite{yang+2022neurips}. 
Unlike methods such as LFTL \cite{lyu+2024eccv} and RLD \cite{song+2024eccv}, our approach requires no manual intervention during adaptation, thereby offering greater flexibility and reproducibility.
While methods that employ less-biased feature extractors \cite{zhang+2023iccv, diamant+2024eccv, tejero+2024eccv} are unable to directly utilize their classification layers for adaptation, our approach seamlessly adapts to target label spaces, thanks to the zero-shot classification capability of ViL models.
Existing SFDA methods that utilize ViL models are closely related to ours. 
However, unlike POUF \cite{tanwisuth+2023icml} and ReCLIP \cite{xuefeng+2024wacv}, which directly adapt CLIP models to the target domain, our approach can integrate any adaptation model ({\it e.g.}, ResNet \cite{he+2016cvpr}), yielding lightweight target models that are more suitable for deployment in resource-constrained environments.
Although Co-learn++ \cite{zhang+2024ijcv} can be used for pseudo-labeling within a variety of SFDA methods, our focus is on enhancing AaD and our approach, ViLAaD, deeply embeds ViL models into its core design. 
Additionally, the alternating optimization framework proposed in DIFO \cite{tang+2024cvpr} can be incorporated into our method. 
As shown in \S\ref{sec:eval:vilaadpp}, the resulting variant, ViLAaD++, achieves state-of-the-art performance across a range of evaluation settings.
\subsection{Vision-and-Language (ViL) Model}
\label{sec:related:vil}
ViL models establish alignment between visual and textual embeddings, enabling robust cross-modal understanding \cite{zhang+2024tpami}.
A key property of ViL models is their zero-shot classification (ZSC) capability: by computing cosine similarities between an image embedding and text embeddings corresponding to different target classes 
({\it e.g.}, embeddings of {\texttt{`a photo of a [CLS]'}}),
classification logits can be directly obtained for any given set of classes.
In this work, we leverage this ViL-based ZSC mechanism to facilitate target model adaptation in the AaD framework.
\par
Recent ViL models such as LLaVA \cite{liu+2023neurips}, InternVL \cite{chen+2024cvpr}, and Qwen-VL \cite{bai+2023arxiv} incorporate large language models (LLMs) to handle complex multimodal tasks. 
However, our focus is on ViL models that extract image and text embeddings using separate modules, without relying on joint processing or LLMs.
Specifically, we employ CLIP \cite{radford+2021arxiv}, ALBEF \cite{li+2021neurips}, and BLIP \cite{li+2022icml} in our implementation.
Although ALBEF and BLIP incorporate feature entanglement mechanisms to enhance cross-modal alignment, we do not utilize these modules in our setup.
\par
While ViL models have been applied to various domain adaptation scenarios \cite{zhu+2024arxiv,lai+2023iccv,singha+2023iccv,du+2024cvpr,westfechtel+2023arxiv,li+2024cvpr,chen+2024arxiv,vesdapunt+2024eccv}, their use in SFDA remains relatively unexplored.
Our approach, inspired by recent efforts \cite{tanwisuth+2023icml,xuefeng+2024wacv,tang+2024cvpr,zhang+2024ijcv,tang+2025iclr}, represents one of the earliest attempts to employ ViL models for addressing the SFDA problem.
\section{Proposed Method}
\label{sec:proposed}
In the standard Source-Free Domain Adaptation (SFDA) setting, we are given a pretrained model $\theta_s$ trained on a labeled source dataset with $|C_s|$ classes, and an unlabeled target dataset $\mathcal{D}_t$ consisting of $N_t$ samples drawn from $|C_t|$ classes.
$C_s$ and $C_t$ denote the source and target class sets, respectively.
The goal of SFDA is to adapt the source model $\theta_s$ to the target domain without access to the source data. The resulting adapted model is denoted as $\theta_t$.
We decompose both the source and target models into two components: a feature extractor $f$ and a classifier $g$. For a given sample $x_i$, the feature extractor produces a representation $z_i = f(x_i) \in \mathbb{R}^h$, where $h$ is the feature dimension. 
The classifier then outputs a probability distribution $p_i = \sigma(g(z_i)) \in \mathbb{R}^{|C|}$, where $\sigma$ denotes the softmax function and $|C|$ is the number of classes.
We consider three SFDA scenarios based on the relationship between $C_s$ and $C_t$:
\begin{itemize}
\item Closed-set SFDA (C-SFDA): $C_s = C_t$
\item Partial-set SFDA (P-SFDA): $C_s \supset C_t$  
\item Open-set SFDA (O-SFDA): $C_s \subset C_t$
\end{itemize}
\par
In addition to the source model, our proposed method assumes access to a Vision-and-Language (ViL) model $\theta_v$.
As discussed in \S\ref{sec:related:vil}, we use CLIP \cite{radford+2021arxiv}, ALBEF \cite{li+2021neurips}, and BLIP \cite{li+2022icml} as instantiations of $\theta_v$ in our experiments.
Given an appropriate text prompt $r$ 
({\it e.g.}, {\texttt{`a photo of a [CLS]'}},
where {\texttt{[CLS]}} is a class name from the target domain),
the ViL model can produce a classifier output $q_i \in \mathbb{R}^{|C|}$ by computing the softmax over the cosine similarities between the image encoder's output for $x_i$ and the text encoder's outputs for each class prompt.
\par
In this section, we first review the Attracting and Dispersing (AaD) \cite{yang+2022neurips} in \S\ref{sec:proposed:aad}.
Then, building upon AaD, we introduce ViLAaD in \S\ref{sec:proposed:vilaad}, which enhances AaD with ViL-based supervision. 
Furthermore, in \S\ref{sec:proposed:vilaadpp}, we introduce ViLAaD++, an extension of ViLAaD that enhances SFDA performance through joint optimization of a learnable ViL prompt and the target model, along with additional adaptation objectives.
\subsection{Attracting and Dispersing (AaD) \cite{yang+2022neurips}}
\label{sec:proposed:aad}
Building on the assumption that the source-pretrained model offers a strong foundation for target adaptation, AaD \cite{yang+2022neurips} tackles the SFDA problem by attracting predictions of features that are close in the target model’s feature space, and dispersing predictions of features that are farther apart.
To quantify prediction similarity, AaD defines the following probability:
\begin{gather}
a_{ij} = \frac{e^{p_{i}^\mathsf{T}p_{j}}}{\sum^{N_{t}}_{k=1} e^{p_{i}^\mathsf{T}p_{k}}},
\end{gather}
which can be interpreted as the probability that sample $x_i$ shares a similar prediction with sample $x_j$ within the target dataset $\mathcal{D}_t$.
For each $x_i$, two sets are constructed:
(1) the close neighbor set $\mathcal{C}_i$, which contains the indices of the $K$-nearest neighbors of $x_i$ in the feature space, and
(2) the complementary set $\bar{\mathcal{C}}_i$, which includes the indices of all samples in $\mathcal{D}_t$ that are not in $\mathcal{C}_i$.
Neighborhood relationships are determined using cosine similarity between features.
Note that the index $i$ itself is included in its own neighbor set $\mathcal{C}_i$.
With $a_{ij}$, $\mathcal{C}_i$, and $\bar{\mathcal{C}}_i$ defined, AaD formulates two likelihood functions to model the attraction and dispersion objectives:
\begin{gather}
A( \mathcal{C}_{i} | \theta ) = \prod_{j \in \mathcal{C}_{i}, j \neq i} a_{ij} = \prod_{j \in \mathcal{C}_{i}, j \neq i} \frac{e^{p_{i}^\mathsf{T}p_{j}}}{\sum^{N_t}_{k=1} e^{p_{i}^{\mathsf{T}} p_{k}}}, \nonumber \\ 
A( \bar{\mathcal{C}}_{i} | \theta ) = \prod_{m \in \bar{\mathcal{C}}_{i}} a_{im} = \prod_{m \in \bar{\mathcal{C}}_{i} } \frac{e^{p_{i}^\mathsf{T}p_{m}}}{\sum^{N_t}_{k=1} e^{p_{i}^{\mathsf{T}} p_{k}}}.  
\label{eq:aad:a}
\end{gather}
Here, $\theta$ denotes the target model being adapted (omitted in subsequent equations for brevity). 
AaD assumes that each $x_i$ should produce more similar predictions to its close neighbors $\mathcal{C}_i$ than to the remaining samples $\bar{\mathcal{C}}_i$.
To enforce this behavior, AaD defines the following negative log-likelihood objective for target adaptation:
\begin{gather}
\tilde{L}^{\text{AaD}} = \mathbb{E} [ \tilde{L}^{\text{AaD}}_{i} ( \mathcal{C}_{i}, \bar{\mathcal{C}}_{i} ) ], \:\: {\text{with}} \: \tilde{L}^{\text{AaD}}_{i} ( \mathcal{C}_{i}, \bar{\mathcal{C}}_{i} ) = - \log \frac{ A(\mathcal{C}_{i}) }{ A( \bar{\mathcal{C}}_{i} ) }.
\label{eq:aad:loss1}
\end{gather}
Unfortunately, minimizing Equation~\eqref{eq:aad:loss1} requires access to all target samples in order to compute $A(\bar{\mathcal{C}}_i)$, which is impractical in real-world settings due to computational and memory constraints. 
To address this issue, \cite{yang+2022neurips} derives an upper bound of Equation~\eqref{eq:aad:loss1}, leading to the following simplified and more tractable objective, denoted as $L^{\text{AaD}}$:
\begin{multline}
L^{\text{AaD}} = \mathbb{E} [ L^{\text{AaD}}_{i} ( \mathcal{C}_{i}, \bar{\mathcal{C}}_{i} ) ], {\text{with}} \\
L^{\text{AaD}}_{i} ( \mathcal{C}_{i}, \bar{\mathcal{C}}_{i} ) = - \sum_{j \in \mathcal{C}_{i}, j \neq i} p_{i}^{\mathsf{T}} p_{j} + \lambda \sum_{m \in \bar{\mathcal{C}}_{i} } p_{i}^{\mathsf{T}} p_{m}.
\label{eq:aad:loss2}
\end{multline}
The first term of $L^{\text{AaD}}_{i}(\mathcal{C}_i, \bar{\mathcal{C}}_i)$ encourages prediction consistency among local neighbors, promoting alignment of semantically similar features. 
In contrast, the second term aims to disperse predictions for potentially dissimilar features. 
Notably, in the AaD implementation, this second term is approximated by computing the average dissimilarity between $p_i$ and all other predictions within the current mini-batch.
\begin{algorithm}[t]
\caption{ViLAaD}
\label{alg:vilaad}
\textbf{Input}: Pre-trained source model $\theta_{s}$, Target dataset $\mathcal{D}^{t}$, ViL model $\theta_{v}$, Prompt $r$, \# epochs $T$, \# iterations per epoch $M$ \\
\textbf{Output}: Target model $\theta_{t}$
\begin{algorithmic}[1] 
\STATE Set $\theta_{t} \leftarrow \theta_{s}$ and $r \leftarrow$ {\texttt{`a photo of a [CLS]'}}
\STATE Initialize memory banks $\mathcal{B}_{z}$ (features produced by $\theta_{t}$), $\mathcal{B}_{p}$ (predictions by $\theta_{t}$) and $\mathcal{B}_{q}$ (predictions by $\theta_{v}$)
\FOR {$t = 1:T$}
\FOR {$m = 1:M$} 
\STATE Sample a mini-batch from $\mathcal{D}^{t}$ and update $\mathcal{B}_{z}$, $\mathcal{B}_{p}$ with current features and predictions
\STATE For each feature $z_{i}$ in the mini-batch, retrieve $K$-nearest neighbors from $\mathcal{B}_{z}$ to construct the close neighbor set $\mathcal{C}_{i}$
\STATE Retrieve ViL predictions that corresponds to the indices in $\mathcal{C}_{i}$ from $\mathcal{B}_{q}$
\STATE Update $\theta_{t}$ by minimizing Eq. \eqref{eq:vilaad:loss}.
\ENDFOR
\ENDFOR
\RETURN $\theta_{t}$.
\end{algorithmic}
\end{algorithm}
\subsection{ViLAaD: ViL-enhanced AaD}
\label{sec:proposed:vilaad}
Here, we aim to enhance the AaD framework by incorporating a ViL model $\theta_v$. 
To this end, we introduce an additional assumption: the ViL model provides a strong (and potentially superior) initialization for target adaptation. 
Intuitively, under this assumption, the predictions of the target model for local neighbors in its feature space should align closely with the corresponding predictions made by the ViL model.
To formalize this intuition, we define $b_{ij}$ as follows:
\begin{gather}
b_{ij} = \frac{e^{p_{i}^\mathsf{T} q_{j}}}{\sum^{N_{t}}_{k=1} e^{p_{i}^\mathsf{T} q_{k}}},
\end{gather}
\begin{algorithm}[t]
\caption{ViLAaD++}
\label{alg:vilaadpp}
\textbf{Input}: Pre-trained source model $\theta_{s}$, Target dataset $\mathcal{D}^{t}$, ViL model $\theta_{v}$, Learnable prompt $r$, \# epochs $T$, \# iterations per epoch $M$ \\
\textbf{Output}: Target model $\theta_{t}$
\begin{algorithmic}[1] 
\STATE Set $\theta_{t} \leftarrow \theta_{s}$ and $r \leftarrow$ {\texttt{`a photo of a [CLS]'}}
\STATE Initialize memory banks $\mathcal{B}_{z}$ (features produced by $\theta_{t}$), $\mathcal{B}_{p}$ (predictions by $\theta_{t}$) and $\mathcal{B}_{q}$ (predictions by $\theta_{v}$)
\FOR {$t = 1:T$}
\STATE \texttt{\% loop to update the learnable prompt}
\FOR {$m = 1:M$} 
\STATE Update $r$ by minimizing Equation \eqref{eq:vilaadpp:prompt} and update $\mathcal{B}_{q}$.
\ENDFOR
\STATE \texttt{\% loop to update the target model}
\FOR {$m = 1:M$} 
\STATE Sample a mini-batch from $\mathcal{D}^{t}$ and update $\mathcal{B}_{z}$, $\mathcal{B}_{p}$ with current features and predictions
\STATE Retrieve ViL predictions that corresponds to the indices in $\mathcal{C}_{i}$ from $\mathcal{B}_{q}$
\STATE Update $\theta_{t}$ by minimizing Eq. \eqref{eq:vilaadpp:all}.
\ENDFOR
\ENDFOR
\RETURN $\theta_{t}$.
\end{algorithmic}
\end{algorithm}
which can be interpreted as the probability that the prediction of a sample $x_i$ by the adapted model ({\it i.e.}, $p_i$) is similar to the prediction of another sample $x_j$ by the ViL model ({\it i.e.}, $q_j$).
Using the same $\mathcal{C}_i$ and $\bar{\mathcal{C}}_i$ defined in \S\ref{sec:proposed:aad}, we introduce the following likelihood functions, analogous to those in Equation~\eqref{eq:aad:a}:
\begin{gather}
B( \mathcal{C}_{i} | \theta ) = \prod_{j \in \mathcal{C}_{i}} b_{ij} = \prod_{j \in \mathcal{C}_{i}} \frac{e^{p_{i}^\mathsf{T}q_{j}}}{\sum^{N_t}_{k=1} e^{p_{i}^{\mathsf{T}} q_{k}}}, \nonumber \\ 
B( \bar{\mathcal{C}}_{i} | \theta ) = \prod_{m \in \bar{\mathcal{C}}_{i}} b_{im} = \prod_{m \in \bar{\mathcal{C}}_{i} } \frac{e^{p_{i}^\mathsf{T}q_{m}}}{\sum^{N_t}_{k=1} e^{p_{i}^{\mathsf{T}} q_{k}}}.
\label{eq:vilaad:b}
\end{gather}
ViLAaD is based on the assumption that the target model’s prediction for $x_i$ should be more similar to the ViL model’s predictions for its close neighbors $\mathcal{C}_i$ than for its distant samples $\bar{\mathcal{C}}_i$.
Based on this assumption, we define the following negative log-likelihood as the ViLAaD objective:
\begin{multline}
\tilde{L}^{\text{ViLAaD}} = \mathbb{E} [ \tilde{L}^{\text{ViLAaD}}_{i} ( \mathcal{C}_{i}, \bar{\mathcal{C}}_{i} ) ], \:\: {\text{with}} \\ 
\tilde{L}^{\text{ViLAaD}}_{i} ( \mathcal{C}_{i}, \bar{\mathcal{C}}_{i} ) = - \log \frac{ B(\mathcal{C}_{i}) }{ B( \bar{\mathcal{C}}_{i} ) }
\label{eq:vilaad:obj}
\end{multline}
However, directly minimizing Equation~\eqref{eq:vilaad:obj} is computationally infeasible. To address this, we derive an upper bound on the objective:
\begin{multline}
\tilde{L}^{\text{ViLAaD}}_{i} ( \mathcal{C}_{i}, \bar{\mathcal{C}}_{i} ) = - \log \frac{ B(\mathcal{C}_{i}) }{ B( \bar{\mathcal{C}}_{i} ) } \\ 
= - \sum_{j \in \mathcal{C}_{i}} \left[ p_{i}^\mathsf{T} q_{j} - \log \left( \sum^{N_t}_{k=1} \exp \left( p_{i}^\mathsf{T} q_{k} \right) \right) \right] \\ 
+ \sum_{m \in \bar{\mathcal{C}}_{i}} \left[ p_{i}^\mathsf{T} q_{m} - \log \left( \sum^{N_t}_{k=1} \exp \left( p_{i}^\mathsf{T} q_{k} \right) \right) \right] \\
= - \sum_{j \in \mathcal{C}_{i}} p_{i}^\mathsf{T} q_{j} + \sum_{m \in \bar{\mathcal{C}}_{i}} p_{i}^\mathsf{T} q_{m} \\ 
+ \left( | \mathcal{C}_{i} | - | \bar{\mathcal{C}}_{i} | \right) \underbrace{ \log \left( \sum^{N_t}_{k=1} \exp \left( p_{i}^\mathsf{T} q_{k} \right) \right) }_{\left( * \right)}.
\label{eq:vilaad:trans}
\end{multline}
To simplify the term (*), we apply Jensen’s inequality:
\begin{multline}
(*) = \log \left( \sum^{N_t}_{k=1} \exp \left( p_{i}^\mathsf{T} q_{k} \right) \right) \leq \sum_{k=1}^{N_t} \frac{1}{N_t} p_{i}^\mathsf{T} q_{k} + \log N_t \\
\simeq \sum_{k=1}^{N_t} \frac{1}{N_t} p_{i}^\mathsf{T} q_{k} + \log N_t
\label{eq:vilaad:asterisk}
\end{multline}
Notice that $N_{t} = | \mathcal{C}_{i} | + | \bar{\mathcal{C}}_{i} |$.
Since the first term on the right-hand side of Equation~\eqref{eq:vilaad:asterisk} can be decomposed into the sum of the inner products over $\mathcal{C}_i$ and $\bar{\mathcal{C}}_i$, the upper bound of $\tilde{L}^{\text{ViLAaD}}$ simplifies to:
\begin{multline}
L^{\text{ViLAaD}} = \mathbb{E} [ L^{\text{ViLAaD}}_{i} ( \mathcal{C}_{i}, \bar{\mathcal{C}}_{i} ) ], \:\: {\text{with}} \\
L^{\text{ViLAaD}}_{i} ( \mathcal{C}_{i}, \bar{\mathcal{C}}_{i} ) = - \sum_{j \in \mathcal{C}_{i}} p_{i}^{\mathsf{T}} q_{j} + \lambda \sum_{m \in \bar{\mathcal{C}}_{i} } p_{i}^{\mathsf{T}} q_{m},
\label{eq:vilaad:loss}
\end{multline}
where $\lambda$ is a hyperparameter.
The ViLAaD algorithm, which minimizes Equation~\eqref{eq:vilaad:loss} to adapt the source model to the target dataset, is outlined in Algorithm~\ref{alg:vilaad}.
Inspired by previous SFDA methods \cite{liang+2020icml,yang+2022neurips,tang+2024cvpr}, we leverage three memory banks to improve training efficiency:
$\mathcal{B}_z$ and $\mathcal{B}_p$ store the features and predictions produced by the adapted model, while $\mathcal{B}_q$ holds the predictions generated by the ViL model.
Note that $\mathcal{B}_z$ and $\mathcal{B}_p$ are dynamically updated throughout training, whereas $\mathcal{B}_q$ remains fixed.
Similar to the approximation used in Equation~\eqref{eq:aad:loss2}, the second term in Equation~\eqref{eq:vilaad:loss} is estimated using the average dissimilarity between $p_i$ and all ViL model predictions within the current mini-batch.
\begin{table*}[]
\centering
\begin{subtable}{0.36\linewidth}
\centering
\scalebox{0.85}{
\begin{tabular}{@{}l|cccccc|c@{}}
\toprule
Method & AD & AW & DA & DW & WA & WD & Avg. \\ \midrule
Source & 79.3 & 76.4 & 59.9 & 95.5 & 61.4 & 98.8 & 78.5      \\ 
ZSC  & 83.1 & 80.5 & 76.3 & 80.5 & 76.3 & 83.1 & 80.0      \\ 
AaD \cite{yang+2022neurips} & 96.4 & 92.1 & 75.0 & \textbf{99.1} & 76.5 & \textbf{100.} & 89.9      \\
\midrule
\rowcolor{red!15}
ViLAaD & \textbf{97.6} & \textbf{93.8} & \textbf{79.6} & 96.7 & \textbf{79.7} & \textbf{100.} & \textbf{91.2}      \\
\bottomrule
\end{tabular}
}
\caption{Office-31}
\end{subtable}
\begin{subtable}{0.54\linewidth}
\centering
\scalebox{0.85}{
\begin{tabular}{@{}cccccccccccc|c@{}}
\toprule
AC & AP & AR & CA & CP & CR & PA & PC & PR & RA & RC & RP & Avg. \\ \midrule
43.7 & 67.0 & 73.9 & 49.9 & 60.1 & 62.5 & 51.7 & 40.9 & 72.6 & 64.2 & 46.3 & 78.1 & 59.2     \\ 
61.2 & 86.0 & 86.3 & 76.9 & 86.0 & 86.3 & 76.9 & 61.2 & 86.3 & 76.9 & 61.2 & 86.0 & 77.6 \\
59.3 & 79.3 & 82.1 & 68.9 & 79.8 & 79.5 & 67.2 & 57.4 & 83.1 & 72.1 & 58.5 & 85.4 & 72.7     \\
\midrule
\rowcolor{red!15}
\textbf{66.4} & \textbf{86.8} & \textbf{87.5} & \textbf{77.2} & \textbf{88.1} & \textbf{87.1} & \textbf{77.1} & \textbf{65.9} & \textbf{88.5} & \textbf{78.7} & \textbf{68.0} & \textbf{89.1} & \textbf{80.0} \\
\bottomrule
\end{tabular}
}
\caption{Office-Home}
\end{subtable}
\begin{subtable}{0.09\linewidth}
\centering
\scalebox{0.85}{
\begin{tabular}{@{}c@{}}
\toprule
Per-class \\ \midrule
49.2     \\ 
86.2 \\ 
88.0 \\
\midrule
\rowcolor{red!15}
\textbf{88.8}     \\
\bottomrule
\end{tabular}
}
\caption{VisDA-C}
\end{subtable}
\caption{
ViLAaD vs. baselines. 
The ``Source'' method applies each source model without adaptation, while ``ZSC'' represents the zero-shot classification performance using a Vision-and-Language (ViL) model. For both ZSC and ViLAaD, we utilize CLIP-ViT-B/32 \cite{radford+2021arxiv} as the ViL model. 
The best results are highlighted in bold.
}
\label{tab:eval:vilaad:c-b32}
\end{table*}
\begin{figure*}[t]
\centering
\includegraphics[width=1.0\textwidth, page=1]{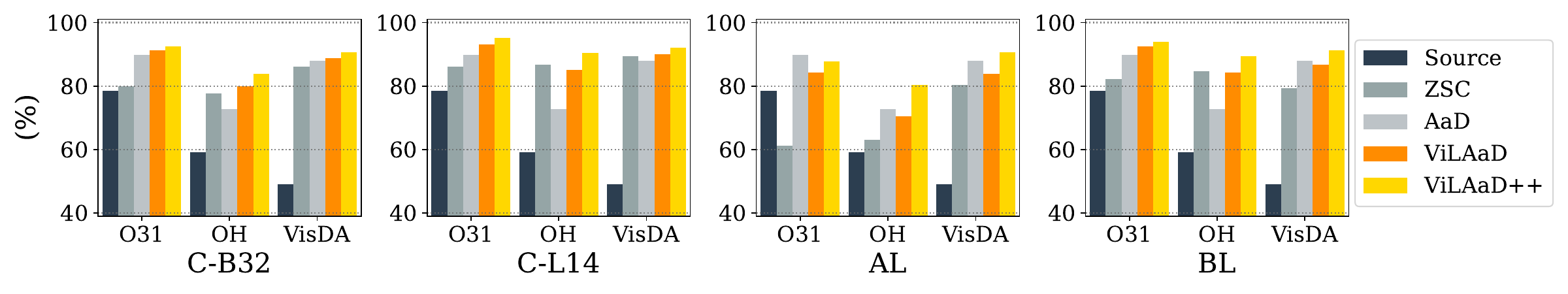}
\caption{
ViLAaD / ViLAaD++ vs. baselines with different ViL models {\it i.e.}, CLIP-ViT-B/32 (C-B32) \cite{radford+2021arxiv}, CLIP-ViT-L/14 (C-L14) \cite{radford+2021arxiv}, ALBEF (AL) \cite{li+2021neurips} and BLIP (BL) \cite{li+2022icml} on the Offce-31 (O31), Offiece-Home (OH) and VisDA-C (VisDA) datasets.
}
\label{fig:eval:vilaad-vils}
\end{figure*}
\begin{figure*}[htbp]
\begin{tabular}{ccc}
\begin{minipage}[t]{0.31\textwidth}
\centering
\includegraphics[keepaspectratio, width=\textwidth]{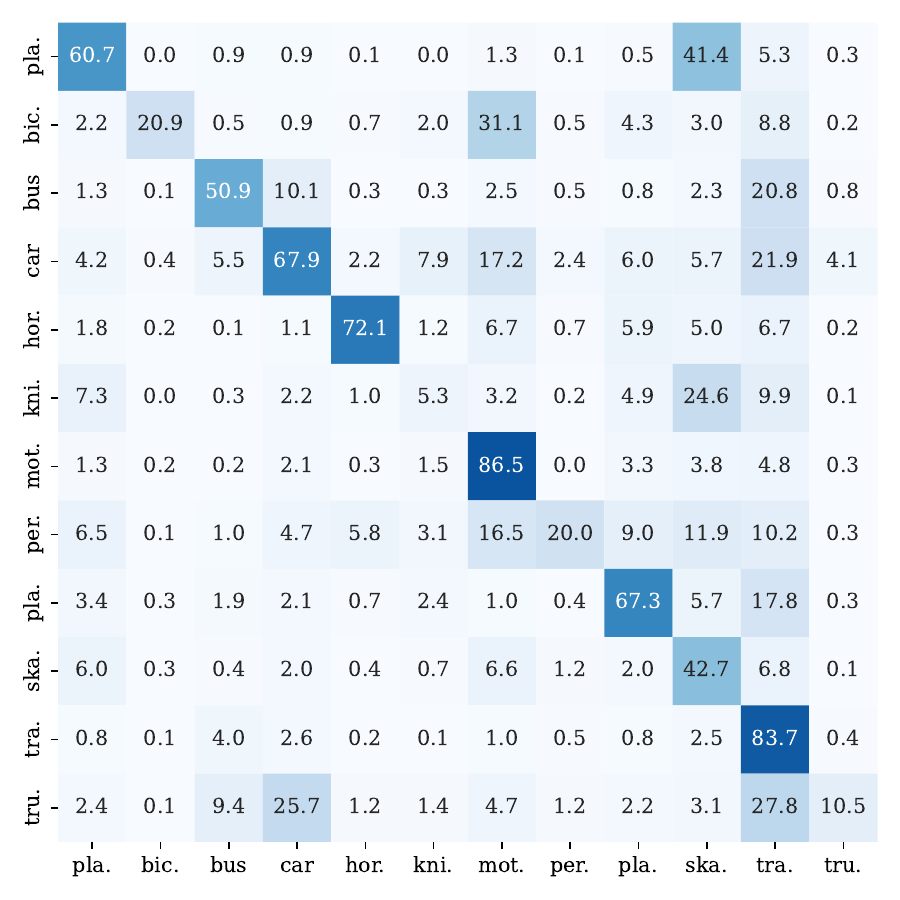}
\subcaption{Source}
\label{fig:eval:conf_mat:conf_src}
\end{minipage} & 
\begin{minipage}[t]{0.31\textwidth}
\centering
\includegraphics[keepaspectratio, width=\textwidth]{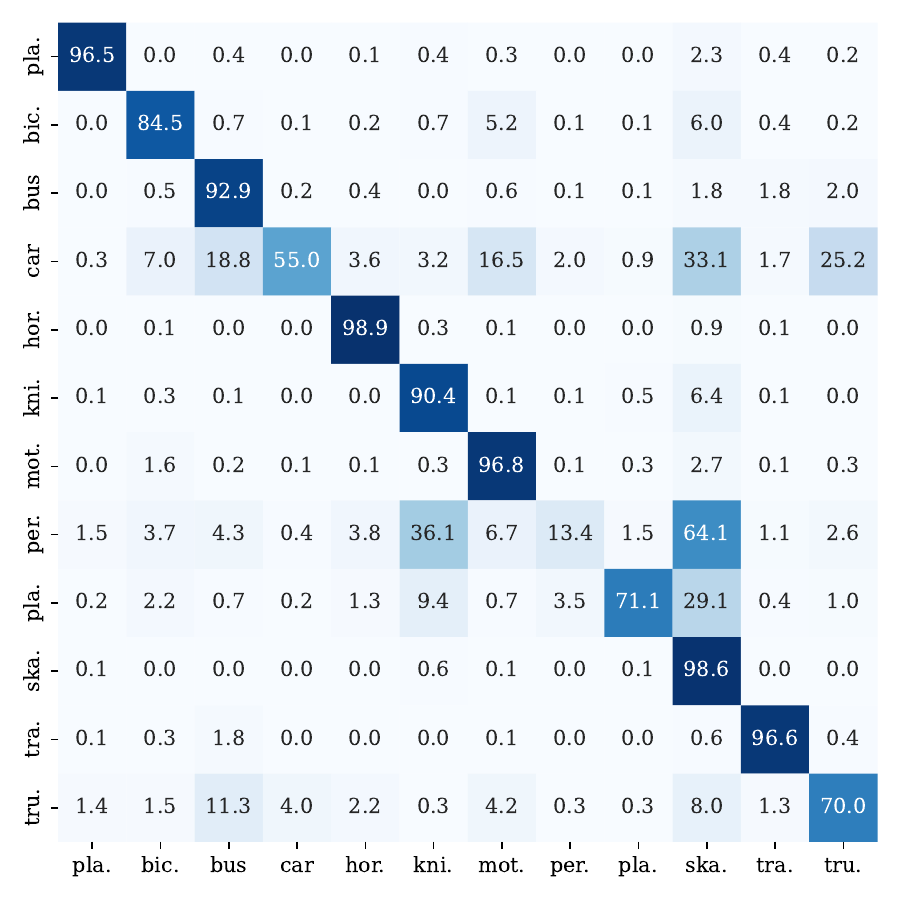}
\subcaption{ZSC w/ ALBEF}
\label{fig:eval:conf_mat:conf_albef}
\end{minipage} &
\begin{minipage}[t]{0.31\textwidth}
\centering
\includegraphics[keepaspectratio, width=\textwidth]{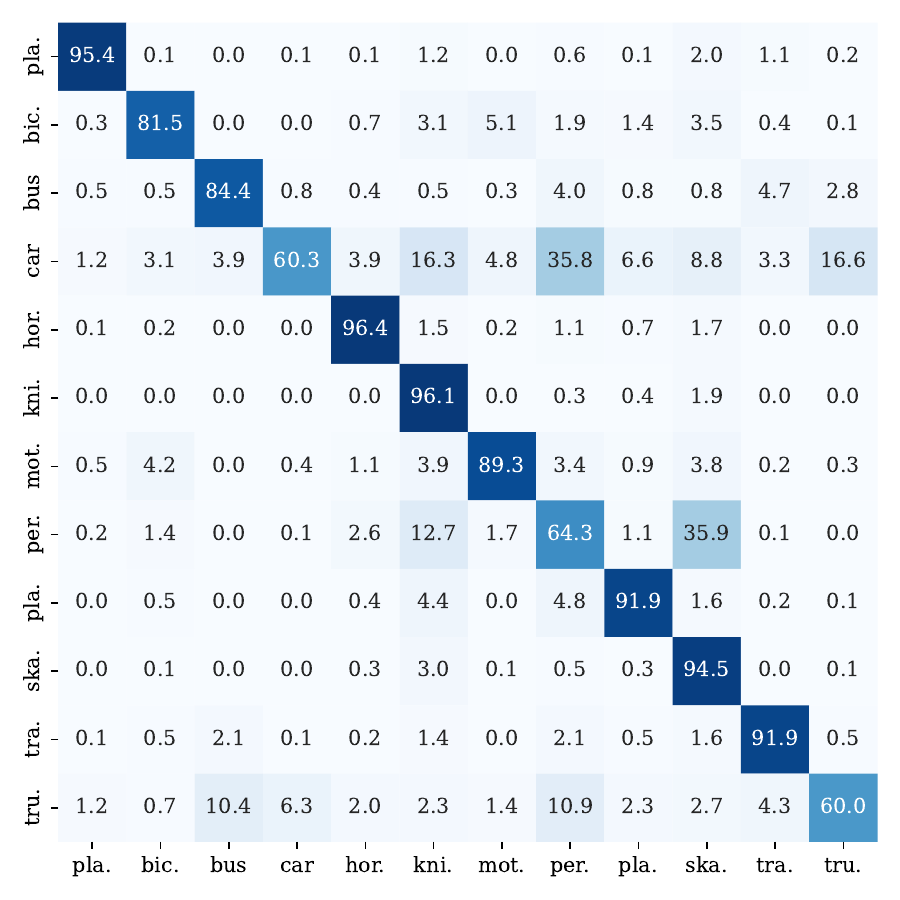}
\subcaption{ViLAaD w/ ALBEF}
\label{fig:eval:conf_mat:conf_blip}
\end{minipage} \\
\end{tabular}
\caption{
Confusion matrices for (a) the source model's predictions, (b) zero-shot classification using ALBEF \cite{li+2021neurips}, and ViLAaD with ALBEF on the VisDA-C dataset.
For the ``car'' class, ZSC achieves an accuracy of 55.0\%, which is lower than that of Source (67.9\%).
As a result, ViLAaD reaches 60.3\% accuracy, which is below both Source and AaD (76.2\%, see Table~\ref{tab:eval:closed:visda}).
}
\label{fig:eval:conf_mat}
\end{figure*}
\subsection{ViLAaD++}
\label{sec:proposed:vilaadpp}
ViLAaD, introduced in \S\ref{sec:proposed:vilaad}, extends AaD by replacing its original objective ({\it i.e.}, Equation~\eqref{eq:aad:loss2}) with our proposed formulation ({\it i.e.}, Equation~\eqref{eq:vilaad:loss}). 
To further improve SFDA performance, we introduce two key enhancements to the ViLAaD framework: (1) an alternating optimization strategy incorporating ViL prompt tuning, and (2) additional objectives designed to better adapt the target model.
The resulting enhanced framework, referred to as ViLAaD++, is summarized in Algorithm~\ref{alg:vilaadpp}.
These enhancements are detailed ad follows:
\par
\vspace{1mm}
\noindent {\bf Alternating Optimization between ViL Prompt Tuning and Target Model Adaptation.}
Thanks to the flexibility of ViLAaD, it can be seamlessly integrated into the alternating optimization framework originally proposed by DIFO~\cite{tang+2024cvpr}. 
In this framework, the parameters of the ViL model are kept frozen, while a learnable prompt for the ViL text encoder is optimized in an alternating manner alongside the target model.
To optimize the prompt $r$, we adopt the same objective introduced by \cite{tang+2024cvpr}, defined as:
\begin{gather}
L^{\text{pro}} = - \min \mathbb{E}_{x_{i} \in \mathcal{D}^{t}} [ M ( \theta_{t} (x_{i}), \theta_{v} (x_{i}, r)) ],
\label{eq:vilaadpp:prompt}
\end{gather}
where $M$ is the mutual information \cite{ji+2019iccv}. 
As shown in Algorithm~\ref{alg:vilaadpp}, ViL prompt tuning (Lines 5-7) and target model adaptation (Lines 9-13) are performed sequentially in each epoch. 
In contrast to ViLAaD, ViLAaD++ updates all memory banks including $\mathcal{B}_{z}$, $\mathcal{B}_{p}$ and $\mathcal{B}_{q}$ throughout the process.
\par
\vspace{1mm}
\noindent {\bf{Additional Objectives for Target Model Adaptation.}}
In addition to the ViLAaD objective defined in Equation~\eqref{eq:vilaad:loss}, we incorporate two conventional objectives to further enhance model adaptation.
The first is a cross-entropy loss, $L^{\text{cls}}$, computed between the model’s predictions and pseudo labels. 
For a target sample $x_i$, the pseudo label $\bar{y}_{i}$ (represented as a one-hot vector) is computed as:
$\bar{y}_{i} = \omega \cdot {\text{one\_hot}} (p_{i}) + ( 1- \omega ) \cdot {\text{one\_hot}} (q_{i})$, 
where ${\text{one\_hot}}(p)$ denotes a function that converts a softmax output $p$ into a one-hot vector. 
Here, $p_i$ and $q_i$ are the most recent predictions made by the target model and the ViL model, respectively. 
The weight $\omega$ is sampled from an exponential distribution with the scale parameter $0.1$.
\par
The second objective is a KL divergence loss, $L^{\text{div}}$, between the empirical label distribution predicted by the target model and a uniform distribution ({\it i.e.}, $\frac{1}{|C|}$ for $|C|$ classes). 
This regularization encourages balanced predictions across all classes and mitigates mode collapse, where the model may otherwise assign most samples to a few dominant classes \cite{liang+2020icml,yang+2021neurips}.
\vspace{1mm}
\par
Finally, the overall objective for target model adaptation in ViLAaD++ is defined as follows:
\begin{gather}
L^{\text{ViLAaD++}} = L^{\text{ViLAaD}} + \lambda^{\text{cls}} L^{\text{cls}} + \lambda^{\text{div}} L^{\text{div}},
\label{eq:vilaadpp:all}
\end{gather}
where $\lambda^{\text{cls}}$ and $\lambda^{\text{div}}$ are both hyperparameters.
\section{Evaluation}
\label{sec:eval}
We begin by outlining our evaluation setup, including datasets (§\ref{sec:eval:dataset}), evaluation metrics (§\ref{sec:eval:metric}), models (§\ref{sec:eval:model}), and implementation details (§\ref{sec:eval:imple}).
Using this setup, we compare ViLAaD with AaD \cite{yang+2022neurips} and zero-shot classification (ZSC) using ViL models in §\ref{sec:eval:vilaad}, to demonstrate the effectiveness of our proposed objective.
We then evaluate ViLAaD++ against state-of-the-art SFDA methods in §\ref{sec:eval:vilaadpp}, showcasing the performance of our approach.
Further analysis of both ViLAaD and ViLAaD++ is provided in §\ref{sec:eval:analysis}.
Note that all the above evaluations are conducted in the Closed-set SFDA (C-SFDA) setting.
Finally, in §\ref{sec:eval:po}, we report the results of ViLAaD++ under Partial-set SFDA (P-SFDA) and Open-set SFDA scenarios.
\subsection{Datasets}
\label{sec:eval:dataset}
In this paper, we evaluate our approach using the following four datasets:
\par
\vspace{1mm}
\noindent {\bf{Office-31 (O31)}}
\cite{saenko+eccv2010} is a small-scale benchmark comprising three domains: Amazon (A), DSLR (D), and Webcam (W).
It contains 4,652 real-world object images spanning 31 categories, collected from diverse office environments.
The Amazon domain consists of e-commerce product images, while the DSLR and Webcam domains provide high-resolution and low-resolution images, respectively.
We use this dataset in the C-SFDA scenario.
\par
\vspace{1mm}
\noindent {\bf{Office-Home (OH)}}
\cite{venkateswara+2017cvpr} is a medium-scale benchmark containing 15,000 images across 65 categories, depicting objects commonly found in work and home environments.
It includes four visually diverse domains: Artistic (A), Clip Art (C), Product (P), and Real-world (R) images.
We use this dataset C-SFDA, P-SFDA and O-SFDA scenarios.
Following the protocol in \cite{liang+2020icml}, for P-SFDA, the target domain contains the first 25 categories (alphabetically), while the source domain includes all 65 categories.
In the O-SFDA setting, the source domain consists of the same 25 categories, and the target domain includes all 65.
Categories not present in the source domain are treated as an unknown class during evaluation.
\par
\vspace{1mm}
\noindent {\bf{VisDA-C (VisDA)}}
\cite{peng+2017arxiv,peng+2018cvprw} is a large-scale and challenging benchmark designed for synthetic-to-real domain adaptation.
It consists of 12 object categories and specifically targets the simulation-to-reality domain shift problem.
The source domain contains 152K synthetic images rendered under diverse angles and lighting conditions, while the target domain includes 55K real images cropped from the Microsoft COCO dataset \cite{lin+2014eccv}.
We employ this dataset under the C-SFDA scenario.
\par
\vspace{1mm}
\noindent {\bf{DomainNet-126}}
\cite{peng+2019iccv,saito+2019iccv} is a large-scale benchmark comprising 145K images across 126 categories.
It covers four visually diverse domains: Clipart (C), Painting (P), Real (R), and Sketch (S).
We utilize this dataset in the C-SFDA scenario.
It is worth noting that while some prior works \cite{hwang+2024iclr,xing+2024eccv} evaluate performance on only 7 out of the 12 possible source-target domain pairs, others consider the full set of combinations.
\begin{table*}[htbp]
\centering
\scalebox{1.0}{
\begin{tabular}{@{}l|l|l|cccccc|c@{}}
\toprule
Method & Venue & Aux. & AD & AW & DA & DW & WA & WD & Avg. \\ \midrule
Source & - & - & 79.3 & 76.4 & 59.9 & 95.5 & 61.4 & 98.8 & 78.5      \\ 
\midrule
ZSC & - & ViL / C-B32 & 83.1 & 80.5 & 76.3 & 80.5 & 76.3 & 83.1 & 80.0 \\ 
ZSC & - & ViL / C-L14 & 87.2 & 87.4 & 83.9 & 87.4 & 83.9 & 87.2 & 86.1 \\ 
ZSC & - & ViL / C-L14@336 & 88.8 & 88.4 & 84.4 & 88.4 & 84.4 & 88.8 & 87.2 \\ 
ZSC & - & ViL / AL & 58.4 & 61.9 & 63.4 & 61.9 & 63.4 & 58.4 & 61.2 \\ 
ZSC & - & ViL / BL & 82.5 & 81.8 & 82.2 & 81.8 & 82.2 & 82.5 & 82.2 \\ 
\midrule
SHOT \cite{liang+2020icml} & ICML20 & - & 93.7 & 91.1 & 74.2 & 98.2 & 74.6 & \textbf{100.} & 88.6 \\
Kim {\it et al}. \cite{kim+2021tai} & TAI21 & - & 92.2 & 91.1 & 71.0 & 98.2 & 71.2 & 99.5 & 87.2 \\
A$^{2}$Net \cite{xia+2021iccv} & ICCV21 & - & 94.5 & 94.0 & 76.7 & 99.2 & 76.1 & \textbf{100.} & 90.1 \\
SHOT \cite{liang+2021tpami} & TPAMI21 & - & 93.9 & 90.1 & 75.3 & 98.7 & 75.0 & 99.9 & 88.8 \\
SHOT++ \cite{liang+2021tpami} & TPAMI21 & - & 94.3 & 90.4 & 76.2 & 98.7 & 75.8 & 99.9 & 89.2 \\
CPGA \cite{qiu+2021ijcai} & IJCAI21 & - & 94.4 & 94.1 & 76.0 & 98.4 & 76.6 & 99.8 & 89.9 \\
GKD \cite{tang+2021iros} & IROS21 &  - & 94.6 & 91.6 &  75.1 & 98.7 & 75.1 & \textbf{100.} & 89.2      \\
NRC \cite{yang+2021neurips} & NeurIPS21 & - & 96.0 & 90.8 & 75.3 & 99.0 & 75.0 & \textbf{100.} & 89.4      \\
AdaCon \cite{chen+2022cvpr} & CVPR22 &  - & 87.7 & 83.1 & 73.7 & 91.3 & 77.6 & 72.8 & 81.0      \\
SFDA-DE \cite{ding+2022cvpr} & CVPR22 &  - & 96.0 & 94.2 & 76.6 & 98.5 & 75.5 & 99.8 & 90.1      \\
CoWA \cite{lee+2022icml} & ICML22 &  - & 94.4 & 95.2 & 76.2 & 98.5 & 77.6 & 99.8 & 90.3      \\
SCLM \cite{tang+2022nn} & NN22 &  - & 95.8 & 90.0 & 75.5 & 98.9 & 75.5 & 99.8 & 89.4      \\
HCL \cite{huang+2022neurips} & NeurIPS22 &  - & 94.7 & 92.5 & 75.9 & 98.2 & 77.7 & \textbf{100.} & 89.8      \\
AaD \cite{yang+2022neurips} & NeurIPS22 &  - & 96.4 & 92.1 & 75.0 & 99.1 & 76.5 & \textbf{100.} & 89.9      \\
ELR \cite{yi+2023iclr} & ICLR23 &  - & 93.8 & 93.3 & 76.2 & 98.0 & 76.9 & \textbf{100.} & 89.6      \\
PLUE \cite{litrico+2023cvpr} & CVPR23 &  - & 89.2 & 88.4 & 72.8 & 97.1 & 69.6 & 97.9 & 85.8      \\
SF(DA)$^{2}$ \cite{hwang+2024iclr} & ICLR24 & - & 95.8 & 92.1 & 75.7 & 99.0 & 76.8 & 99.8 & 89.9 \\
TPDS \cite{tang+2024ijcv} & IJCV24 &  - & 97.1 & 94.5 & 75.7 & 98.7 & 75.5 & 99.8 & 90.2      \\
SHOT+DPC \cite{xia+2024cvpr} & CVPR24 &  - & 95.9 & 92.6 & 75.4 & 98.6 & 76.2 & \textbf{100.} & 89.8      \\
AaD+DPC \cite{xia+2024cvpr} & CVPR24 &  - & 95.8 & 94.5 & 76.5 & 98.9 & 76.8 & \textbf{100.} & 90.5      \\
Improved SFDA \cite{mitsuzumi+2024cvpr} & CVPR24 &  - & 95.3 & 94.2 & 76.4 & 98.3 & 77.5 & 99.9 & 90.3      \\
HRD++ \cite{xing+2024eccv} & ECCV24 & - & 96.5 & 94.5 & 77.1 & 99.1 & 77.9 & 99.9 & 90.8 \\
\midrule
SHOT w/ Co-learn \cite{zhang+2024ijcv} & IJCV24 & FE / C-L14@336 & 95.6 & 92.8 & 77.4 & 98.7 & 77.5 & \textbf{100.} & 90.4 \\
SHOT++ w/ Co-learn \cite{zhang+2024ijcv} & IJCV24 & FE / C-L14@336 & 95.4 & 95.2 & 78.9 & 98.9 & 78.5 & \textbf{100.} & 91.1 \\
NRC w/ Co-learn \cite{zhang+2024ijcv} & IJCV24 & FE / C-L14@336 & 96.6 & 96.3 & 78.0 & 98.5 & 77.9 & \textbf{100.} & 91.2 \\
AaD w/ Co-learn \cite{zhang+2024ijcv} & IJCV24 & FE / C-L14@336 & 95.2 & 96.2 & 78.5 & 98.6 & 79.7 & \textbf{100.} & 91.4 \\
ZSC w/ Co-learn \cite{zhang+2024ijcv} & IJCV24 & FE / C-L14@336 & 99.2 & \textbf{99.7} & 85.3 & 99.1 & 83.2 & \textbf{100.} & 94.4 \\
SHOT w/ Co-learn++ \cite{zhang+2024ijcv} & IJCV24 & ViL / C-L14@336 & 96.2 & 94.7 & 78.3 & 98.2 & 77.6 & 99.8 & 90.8 \\
SHOT++ w/ Co-learn++ \cite{zhang+2024ijcv} & IJCV24 & ViL / C-L14@336 & 96.8 & 92.3 & 78.3 & 98.4 & 77.8 & 99.8 & 90.6 \\
NRC w/ Co-learn++ \cite{zhang+2024ijcv} & IJCV24 & ViL / C-L14@336 & 96.4 & 95.8 & 78.7 & 98.9 & 78.5 & 99.8 & 91.4 \\
AaD w/ Co-learn++ \cite{zhang+2024ijcv} & IJCV24 & ViL / C-L14@336 & 98.0 & 97.7 & 81.3 & 99.1 & 82.1 & \textbf{100.} & 93.0 \\
ZSC w/ Co-learn++ \cite{zhang+2024ijcv} & IJCV24 & ViL / C-L14@336 & \textbf{99.6} & 99.0 & 86.3 & 99.1 & 84.8 & \textbf{100.} & 94.8 \\
DIFO \cite{tang+2024cvpr} & CVPR24 & ViL / C-B32 & 97.2 & 95.5 & 83.0 & 97.2 & 83.2 & 98.8 & 92.5      \\ 
LFTL \cite{lyu+2024eccv} & ECCV24 & MMA & 98.9 & 99.4 & 87.8 & \textbf{100.} & 86.3 & \textbf{100.} & 95.4 \\
\midrule
\rowcolor{red!15}
ViLAaD++ & - & ViL / C-B32 & 96.6 & 95.1 & 85.1 & 96.0 & 83.7 & 99.0 & 92.6 \\
\rowcolor{red!15}
ViLAaD++ & - & ViL / C-L14 & 99.0 & 98.9 & \textbf{88.3} & 99.3 & 87.8 & \textbf{100.} & \textbf{95.5} \\
\rowcolor{red!15}
ViLAaD++ & - & ViL / C-L14@336 & 99.2 & 98.4 & 88.2 & 99.0 & \textbf{88.0} & \textbf{100.} & \textbf{95.5} \\
\rowcolor{red!15}
ViLAaD++ & - & ViL / AL & 83.5 & 83.9 & 79.6 & 93.8 & 79.0 & 99.2 & 86.5 \\
\rowcolor{red!15}
ViLAaD++ & - & ViL / BL & 96.4 & 97.0 & 86.4 & 97.4 & 86.4 & 99.4 & 93.8 \\
\bottomrule
\end{tabular}
}
\caption{
Closed-set SFDA (C-SFDA) results on Office-31 \cite{saenko+eccv2010}. 
The table is divided into five blocks:
(1) Source model baseline,
(2) Zero-shot classification (ZSC) using ViL models,
(3) SFDA without auxiliary resources,
(4) SFDA leveraging auxiliary resources ({\it e.g.}, feature extractors (FE), minimal manual annotation (MMA), ViL models (ViL)),
(5) ViLAaD++ (proposed).
The best results are highlighted in bold.
}
\label{tab:eval:closed:o31}
\end{table*}
\begin{table*}[htbp]
\centering
\scalebox{0.9}{
\begin{tabular}{@{}l|l|l|cccccccccccc|c@{}}
\toprule
Method & Venue & Aux. & AC & AP & AR & CA & CP & CR & PA & PC & PR & RA & RC & RP & Avg. \\ \midrule
Source & - & - & 43.7 & 67.0 & 73.9 & 49.9 & 60.1 & 62.5 & 51.7 & 40.9 & 72.6 & 64.2 & 46.3 & 78.1 & 59.2     \\ 
\midrule
ZSC & - & ViL / C-B32 & 61.2 & 86.0 & 86.3 & 76.9 & 86.0 & 86.3 & 76.9 & 61.2 & 86.3 & 76.9 & 61.2 & 86.0 & 77.6 \\
ZSC & - & ViL / C-L14 & 74.5 & 92.5 & 92.9 & 87.1 & 92.5 & 92.9 & 87.1 & 74.5 & 92.9 & 87.1 & 74.5 & 92.5 & 86.7 \\
ZSC & - & ViL / C-L14@336 & 75.8 & 93.3 & 93.8 & 88.4 & 93.3 & 93.8 & 88.4 & 75.8 & 93.8 & 88.4 & 75.8 & 93.3 & 87.8 \\
ZSC & - & ViL / AL & 52.9 & 68.2 & 70.4 & 61.1 & 68.2 & 70.4 & 61.1 & 52.9 & 70.4 & 61.1 & 52.9 & 68.2 & 63.2 \\ 
ZSC & - & ViL / BL & 76.8 & 91.1 & 88.3 & 82.3 & 91.1 & 88.3 & 82.3 & 76.8 & 88.3 & 82.3 & 76.8 & 91.1 & 84.6 \\ 
\midrule
Kim {\it et al}. \cite{kim+2021tai} & TAI21 & - & 48.4 & 73.4 & 76.9 & 64.3 & 69.8 & 71.7 & 62.7 & 45.3 & 76.6 & 69.8 & 50.5 & 79.0 & 65.7 \\
A$^{2}$Net \cite{xia+2021iccv} & ICCV21 & - & 58.4 & 79.0 & 82.4 & 67.5 & 79.3 & 78.9 & 68.0 & 56.2 & 82.9 & 74.1 & 60.5 & 85.0 & 72.8 \\
SHOT \cite{liang+2020icml} & ICML20 & - & 56.7 & 77.9 & 80.6 & 68.0 & 78.0 & 79.4 & 67.9 & 54.5 & 82.3 & 74.2 & 58.6 & 84.5 & 71.9 \\
SHOT \cite{liang+2021tpami} & TPAMI21 & - & 57.7 & 79.1 & 81.5 & 67.6 & 77.9 & 77.8 & 68.1 & 55.8 & 82.0 & 72.8 & 59.7 & 84.4 & 72.0 \\
SHOT++ \cite{liang+2021tpami} & TPAMI21 & - & 57.9 & 79.7 & 82.5 & 68.5 & 79.6 & 79.3 & 68.5 & 57.0 & 83.0 & 73.7 & 60.7 & 84.9 & 73.0 \\
CPGA \cite{qiu+2021ijcai} & IJCAI21 & - & 59.3 & 78.1 & 79.8 & 65.4 & 75.5 & 76.4 & 65.7 & 58.0 & 81.0 & 72.0 & 64.4 & 83.3 & 71.6 \\
GKD \cite{tang+2021iros} & IROS21 & - & 56.5 & 78.2 & 81.8 & 68.7 & 78.9 & 79.1 & 67.6 & 54.8 & 82.6 & 74.4 & 58.5 & 84.8 & 72.2 \\
NRC \cite{yang+2021neurips} & NeurIPS21 & - & 57.7 & 80.3 & 82.0 & 68.1 & 79.8 & 78.6 & 65.3 & 56.4 & 83.0 & 71.0 & 58.6 & 85.6 & 72.2     \\
AaD \cite{yang+2022neurips} & NeurIPS22 & - & 59.3 & 79.3 & 82.1 & 68.9 & 79.8 & 79.5 & 67.2 & 57.4 & 83.1 & 72.1 & 58.5 & 85.4 & 72.7     \\
DaC \cite{zhang+2022neurips} & NeurIPS22 & - & 59.1 & 79.5 & 81.2 & 69.3 & 78.9 & 79.2 & 67.4 & 56.4 & 82.4 & 74.0 & 61.4 & 84.4 & 72.8     \\
AdaCon \cite{chen+2022cvpr} & CVPR22 & - & 47.2 & 75.1 & 75.5 & 60.7 & 73.3 & 73.2 & 60.2 & 45.2 & 76.6 & 65.6 & 48.3 & 79.1 & 65.0     \\
SFDA-DE \cite{ding+2022cvpr} & CVPR22 & - & 59.7 & 79.5 & 82.4 & 69.7 & 78.6 & 79.2 & 66.1 & 57.2 & 82.6 & 73.9 & 60.8 & 85.5 & 72.9     \\
CoWA \cite{lee+2022icml} & ICML22 & - & 56.9 & 78.4 & 81.0 & 69.1 & 80.0 & 79.9 & 67.7 & 57.2 & 82.4 & 72.8 & 60.5 & 84.5 & 72.5     \\
SCLM \cite{tang+2022nn} & NN22 & - & 58.2 & 80.3 & 81.5 & 69.3 & 79.0 & 80.7 & 69.0 & 56.8 & 82.7 & 74.7 & 60.6 & 85.0 & 73.0     \\
ELR \cite{yi+2023iclr} & ICLR23 & - & 58.4 & 78.7 & 81.5 & 69.2 & 79.5 & 79.3 & 66.3 & 58.0 & 82.6 & 73.4 & 59.8 & 85.1 & 72.6     \\
TPDS \cite{tang+2024ijcv} & IJCV24 & - & 59.3 & 80.3 & 82.1 & 70.6 & 79.4 & 80.9 & 69.8 & 56.8 & 82.1 & 74.5 & 61.2 & 85.3 & 73.5     \\
SHOT+DPC \cite{xia+2024cvpr} & CVPR24 & - & 59.2 & 79.8 & 82.6 & 68.9 & 79.7 & 79.5 & 68.6 & 56.5 & 82.9 & 73.9 & 61.2 & 85.4 & 73.2     \\
AaD+DPC \cite{xia+2024cvpr} & CVPR24 & - & 59.5 & 80.6 & 82.9 & 69.4 & 79.3 & 80.1 & 67.3 & 57.2 & 83.7 & 73.1 & 58.9 & 84.9 & 73.1     \\
AaD w/DCPL \cite{diamant+2024eccv} & ECCV24 & - & 61.9 & 84.5 & 87.1 & 75.7 & 85.8 & 85.6 & 73.6 & 59.8 & 86.5 & 78.1 & 62.9 & 89.0 & 77.5 \\
SHOT w/DCPL \cite{diamant+2024eccv} & ECCV24 & - & 59.9 & 84.3 & 87.8 & 76.8 & 85.8 & 86.6 & 74.8 & 58.6 & 87.4 & 77.9 & 61.1 & 89.0 & 77.5 \\
SHOT++ w/DCPL \cite{diamant+2024eccv} & ECCV24 & - & 61.2 & 84.3 & 88.0 & 76.7 & 86.1 & 86.6 & 74.2 & 59.3 & 87.6 & 78.2 & 62.4 & 89.5 & 77.8 \\
Improved SFDA \cite{mitsuzumi+2024cvpr} & CVPR24 & - & 60.7 & 78.9 & 82.0 & 69.9 & 79.5 & 79.7 & 67.1 & 58.8 & 82.3 & 74.2 & 61.3 & 86.4 & 73.4 \\
HRD++ \cite{xing+2024eccv} & ECCV24 & - & 63.6 & 83.6 & 85.4 & 71.2 & 83.3 & 83.9 & 70.1 & 60.9 & 85.2 & 77.3 & 64.5 & 88.5 & 76.5  \\
\midrule
SHOT w/ Co-learn \cite{zhang+2024ijcv} & IJCV24 & FE / C-L14@336 & 63.1 & 83.1 & 84.6 & 72.0 & 81.9 & 82.0 & 70.8 & 60.4 & 83.8 & 75.9 & 66.1 & 86.1 & 75.8 \\ 
SHOT++ w/ Co-learn \cite{zhang+2024ijcv} & IJCV24 & FE / C-L14@336 & 63.6 & 83.6 & 84.8 & 71.4 & 81.7 & 81.7 & 70.2 & 58.7 & 84.4 & 76.3 & 66.3 & 86.3 & 75.8 \\ 
NRC w/ Co-learn \cite{zhang+2024ijcv} & IJCV24 & FE / C-L14@336 & 72.2 & 87.6 & 88.4 & 77.8 & 87.3 & 88.3 & 77.9 & 70.7 & 89.4 & 79.9 & 74.2 & 90.7 & 82.0 \\ 
AaD w/ Co-learn \cite{zhang+2024ijcv} & IJCV24 & FE / C-L14@336 & 66.4 & 85.3 & 87.0 & 74.7 & 87.1 & 85.6 & 73.0 & 66.8 & 85.9 & 76.4 & 67.2 & 89.8 & 78.8 \\ 
ZSC w/ Co-learn \cite{zhang+2024ijcv} & IJCV24 & FE / C-L14@336 & 77.2 & 90.4 & 91.0 & 77.1 & 88.1 & 90.0 & 76.6 & 72.5 & 90.1 & 82.0 & 79.6 & 93.0 & 84.0 \\ 
SHOT w/ Co-learn++ \cite{zhang+2024ijcv} & IJCV24 & ViL / C-L14@336 & 62.2 & 83.1 & 84.9 & 71.5 & 81.7 & 81.7 & 70.9 & 61.9 & 84.1 & 75.9 & 65.5 & 86.6 & 75.8 \\ 
SHOT++ w/ Co-learn++ \cite{zhang+2024ijcv} & IJCV24 & ViL / C-L14@336 & 62.5 & 83.5 & 84.5 & 72.7 & 81.5 & 83.2 & 71.1 & 61.3 & 84.2 & 76.6 & 65.9 & 86.4 & 76.1 \\ 
NRC w/ Co-learn++ \cite{zhang+2024ijcv} & IJCV24 & ViL / C-L14@336 & 76.4 & 88.8 & 88.6 & 82.4 & 89.1 & 88.2 & 81.0 & 75.2 & 89.7 & 82.4 & 76.3 & 91.9 & 84.2 \\ 
AaD w/ Co-learn++ \cite{zhang+2024ijcv} & IJCV24 & ViL / C-L14@336 & 71.9 & 88.1 & 86.7 & 79.0 & 88.9 & 86.2 & 75.7 & 70.7 & 87.8 & 79.2 & 72.5 & 90.3 & 81.4 \\ 
ZSC w/ Co-learn++ \cite{zhang+2024ijcv} & IJCV24 & ViL / C-L14@336 & 80.0 & 91.2 & 91.8 & 83.4 & 92.7 & 91.3 & 83.4 & 78.9 & 92.0 & 85.5 & 80.6 & 94.7 & 87.1 \\ 
DIFO \cite{tang+2024cvpr} & CVPR24 & ViL / C-B32 & 70.6 & 90.6 & 88.8 & 82.5 & 90.6 & 88.8 & 80.9 & 70.1 & 88.9 & 83.4 & 70.5 & 91.2 & 83.1     \\ 
RLD w/ NBF \cite{song+2024eccv} & ECCV24 & UF & 62.2 & 81.0 & 79.7 & 68.8 & 85.4 & 78.6 & 67.7 & 61.7 & 79.5 & 69.0 & 64.1 & 88.2 & 73.8 \\
LFTL \cite{lyu+2024eccv} & ECCV24 & MMA & 76.6 & 92.2 & 89.7 & 78.9 & 93.0 & 89.2 & 78.6 & 77.1 & 90.0 & 83.4 & 77.8 & 94.6 & 85.1 \\ 
\midrule
\rowcolor{red!15}
ViLAaD++ & - & ViL / C-B32 & 70.1 & 91.6 & 89.9 & 83.2 & 92.0 & 90.0 & 81.0 & 71.7 & 89.9 & 83.3 & 71.3 & 92.3 & 83.9 \\
\rowcolor{red!15}
ViLAaD++ & - & ViL / C-L14 & 82.2 & 95.3 & 94.3 & \textbf{90.3} & \textbf{95.6} & 94.2 & 89.7 & 82.0 & 94.1 & \textbf{90.4} & 81.6 & 95.6 & 90.4 \\
\rowcolor{red!15}
ViLAaD++ & - & ViL / C-L14@336 & 82.7 & \textbf{95.8} & \textbf{94.5} & 89.7 & 95.4 & \textbf{94.7} & \textbf{90.0} & 82.2 & \textbf{94.2} & 90.3 & 82.5 & \textbf{95.8} & \textbf{90.6} \\
\rowcolor{red!15}
ViLAaD++ & - & ViL / AL & 71.4 & 85.7 & 85.8 & 78.9 & 85.6 & 84.6 & 77.8 & 70.3 & 85.0 & 80.6 & 72.4 & 86.7 & 80.4  \\
\rowcolor{red!15}
ViLAaD++ & - & ViL / BL & \textbf{83.3} & 95.1 & 92.3 & 87.8 & 95.1 & 92.4 & 86.3 & \textbf{82.7} & 92.4 & 87.6 & \textbf{83.2} & 95.3 & 89.5  \\
\bottomrule
\end{tabular}
}
\caption{
C-SFDA results on Office-Home \cite{venkateswara+2017cvpr}. 
The table is divided into five blocks:
(1) Source model baseline,
(2) ZSC using ViL models,
(3) SFDA without auxiliary resources,
(4) SFDA leveraging auxiliary resources ({\it e.g.}, feature extractors (FE), minimal manual annotation (MMA), user feedback (UF), ViL models (ViL)),
(5) ViLAaD++ (proposed).
The best results are highlighted in bold.
}
\label{tab:eval:closed:oh}
\end{table*}
\begin{table*}[htbp]
\centering
\scalebox{0.9}{
\begin{tabular}{@{}l|l|l|cccccccccccc|c@{}}
\toprule
Method & Venue & Aux. & pla. & bic. & bus & car & hor. & kni. & mot. & per. & pla. & ska. & tra. & tru. & Per-class \\ 
\midrule
Source & - & - & 60.7 & 21.7 & 50.8 & 68.5 & 71.8 & 5.4 & 86.4 & 20.2 & 67.1 & 43.3 & 83.3 & 10.6 & 49.2 \\
\midrule
ZSC & - & ViL / C-B32 & 98.4 & 87.1 & 91.3 & 69.0 & 98.3 & 88.2 & 91.7 & 76.1 & 75.4 & 94.3 & 94.5 & 69.9 & 86.2 \\ 
ZSC & - & ViL / C-L14 & \textbf{99.6} & 91.4 & 92.6 & 69.3 & 99.5 & 92.2 & 97.1 & 79.8 & 83.9 & \textbf{99.2} & 97.0 & 71.6 & 89.4 \\ 
ZSC & - & ViL / C-L14@336 & \textbf{99.6} & 92.1 & 92.7 & 70.5 & 99.4 & 92.4 & \textbf{97.4} & 79.4 & 85.2 & 98.9 & \textbf{97.1} & 71.6 & 89.7 \\ 
ZSC & - & ViL / AL & 96.5 & 84.5 & 92.9 & 55.0 & 98.9 & 90.4 & 96.8 & 13.4 & 71.1 & 98.6 & 96.6 & 70.0 & 80.4 \\ 
ZSC & - & ViL / BL & 99.4 & 79.5 & \textbf{93.4} & 48.8 & 98.4 & 64.6 & 95.7 & 63.3 & 51.3 & 95.8 & 95.7 & 67.1 & 79.4 \\ 
\midrule
Kim {\it et al}. \cite{kim+2021tai} & TAI21 & - & 86.9 & 81.7 & 84.6 & 63.9 & 93.1 & 91.4 & 86.6 & 71.9 & 84.5 & 58.2 & 74.5 & 42.7 & 76.7 \\
A$^{2}$Net \cite{xia+2021iccv} & ICCV21 & - & 94.0 & 87.8 & 85.6 & 66.8 & 93.7 & 95.1 & 85.8 & 81.2 & 91.6 & 88.2 & 86.5 & 56.0 & 84.3 \\
SHOT \cite{liang+2020icml} & ICML20 & - & 95.0 & 87.4 & 80.9 & 57.6 & 93.9 & 94.1 & 79.4 & 80.4 & 90.9 & 89.8 & 85.8 & 57.5 & 82.7 \\
SHOT \cite{liang+2021tpami} & TPAMI21 & - & 95.8 & 88.2 & 87.2 & 73.7 & 95.2 & 96.4 & 87.9 & 84.5 & 92.5 & 89.3 & 85.7 & 49.1 & 85.5 \\
SHOT++ \cite{liang+2021tpami} & TPAMI21 & - & 97.7 & 88.4 & 90.2 & 86.3 & 97.9 & 98.6 & 92.9 & 84.1 & 97.1 & 92.2 & 93.6 & 28.8 & 87.3 \\
CPGA \cite{qiu+2021ijcai} & IJCAI21 & - & 95.6 & 89.0 & 75.4 & 64.9 & 91.7 & 97.5 & 89.7 & 83.8 & 93.9 & 93.4 & 87.7 & 69.0 & 86.0 \\
NRC \cite{yang+2021neurips} & NeurIPS21 & - & 96.8 & 91.3 & 82.4 & 62.4 & 96.2 & 95.9 & 86.1 & 90.7 & 94.8 & 94.1 & 90.4 & 59.7 & 85.9     \\
GKD \cite{tang+2021iros} & IROS21 & - & 95.3 & 87.6 & 81.7 & 58.1 & 93.9 & 94.0 & 80.0 & 80.0 & 91.2 & 91.0 & 86.9 & 56.1 & 83.0     \\
AaD \cite{yang+2022neurips} & NeurIPS22 & - & 97.4 & 90.5 & 80.8 & 76.2 & 97.3 & 96.1 & 89.8 & 82.9 & 95.5 & 93.0 & 92.0 & 64.7 & 88.0 \\
DaC \cite{zhang+2022neurips} & NeurIPS22 & - & 96.6 & 86.8 & 86.4 & 78.4 & 96.4 & 96.2 & 93.6 & 83.8 & 96.8 & 95.1 & 89.6 & 50.0 & 87.3 \\
AdaCon \cite{chen+2022cvpr} & CVPR22 & - & 97.0 & 84.7 & 84.0 & 77.3 & 96.7 & 93.8 & 91.9 & 84.8 & 94.3 & 93.1 & 94.1 & 49.7 & 86.8     \\
SFDA-DE \cite{ding+2022cvpr} & CVPR22 & - & 95.3 & 91.2 & 77.5 & 72.1 & 95.7 & 97.8 & 85.5 & 86.1 & 95.5 & 93.0 & 86.3 & 61.6 & 86.5     \\
CoWA \cite{lee+2022icml} & ICML22 & - & 96.2 & 89.7 & 83.9 & 73.8 & 96.4 & 97.4 & 89.3 & 86.8 & 94.6 & 92.1 & 88.7 & 53.8 & 86.9     \\
SCLM \cite{tang+2022nn} & NN22 & - & 97.1 & 90.7 & 85.6 & 62.0 & 97.3 & 94.6 & 81.8 & 84.3 & 93.6 & 92.8 & 88.0 & 55.9 & 85.3     \\
ELR \cite{yi+2023iclr} & ICLR23 & - & 97.1 & 89.7 & 82.7 & 62.0 & 96.2 & 97.0 & 87.6 & 81.2 & 93.7 & 94.1 & 90.2 & 58.6 & 85.8     \\
PLUE \cite{litrico+2023cvpr} & CVPR23 & - & 94.4 & 91.7 & 89.0 & 70.5 & 96.6 & 94.9 & 92.2 & 88.8 & 92.9 & 95.3 & 91.4 & 61.6 & 88.3     \\
TPDS \cite{tang+2024ijcv} & IJCV24 & - & 97.6 & 91.5 & 89.7 & 83.4 & 97.5 & 96.3 & 92.2 & 82.4 & 96.0 & 94.1 & 90.9 & 40.4 & 87.6     \\
SF(DA)$^2$ \cite{hwang+2024iclr} & ICLR24 & - & 96.8 & 89.3 & 82.9 & 81.4 & 96.8 & 95.7 & 90.4 & 81.3 & 95.5 & 93.7 & 88.5 & 64.7 & 88.1 \\ 
SHOT+DPC \cite{xia+2024cvpr} & CVPR24 & - & 95.6 & 88.2 & 82.8 & 59.4 & 92.5 & 95.7 & 85.6 & 81.7 & 91.6 & 90.9 & 87.6 & 60.1 & 84.3 \\ 
AaD+DPC \cite{xia+2024cvpr} & CVPR24 & - & 96.5 & 89.3 & 86.5 & 83.2 & 97.4 & 97.3 & 91.8 & 83.7 & 96.4 & 94.8 & 92.1 & 56.2 & 88.8 \\ 
Improved SFDA \cite{mitsuzumi+2024cvpr} & CVPR24 & - & 97.5 & 91.4 & 87.9 & 79.4 & 97.2 & 97.2 & 92.2 & 83.0 & 96.4 & 94.2 & 91.1 & 53.0 & 88.4 \\ 
AaD w/DCPL \cite{diamant+2024eccv} & ECCV24 & - & 97.9 & 92.7 & 83.8 & 77.9 & 97.4 & 97.3 & 90.6 & 83.7 & 96.6 & 96.5 & 92.0 & 66.6 & 89.4 \\
SHOT w/DCPL \cite{diamant+2024eccv} & ECCV24 & - & 97.0 & 89.1 & 83.5 & 63.3 & 95.5 & 97.7 & 88.6 & 81.1 & 93.9 & 95.3 & 90.0 & 65.7 & 86.7 \\
SHOT++ w/DCPL \cite{diamant+2024eccv} & ECCV24 & - & 98.0 & 91.3 & 88.5 & 77.9 & 98.0 & 98.7 & 93.3 & 84.6 & 97.3 & 94.9 & 93.8 & 54.7 & 89.3 \\
SFDA-CDS \cite{tejero+2024eccv} & ECCV24 & - & 91.4 & 83.0 & 83.3 & 75.0 & 98.0 & 82.2 & 92.8 & 85.4 & 95.4 & 90.6 & 87.3 & 41.8 & 83.9 \\ 
\midrule
SHOT w/ Co-learn \cite{zhang+2024ijcv} & IJCV24 & FE / C-L14@336 & 96.3 & 89.8 & 83.8 & 63.0 & 95.6 & 96.7 & 88.4 & 82.1 & 91.7 & 91.4 & 88.6 & 62.2 & 85.8 \\
SHOT++ w/ Co-learn \cite{zhang+2024ijcv} & IJCV24 & FE / C-L14@336 & 97.7 & 91.7 & 89.1 & 83.7 & 98.0 & 97.4 & 90.7 & 84.2 & 97.5 & 94.7 & 94.4 & 39.4 & 88.2 \\
NRC w/ Co-learn \cite{zhang+2024ijcv} & IJCV24 & FE / C-L14@336 & 97.5 & 91.9 & 83.7 & 65.0 & 96.7 & 97.5 & 88.3 & 81.1 & 93.0 & 95.5 & 91.6 & 59.5 & 86.8 \\
AaD w/ Co-learn \cite{zhang+2024ijcv} & IJCV24 & FE / C-L14@336 & 97.5 & 91.4 & 85.4 & 82.4 & 97.3 & 97.8 & 92.3 & 81.7 & 95.7 & 94.3 & 92.5 & 51.7 & 88.3 \\
ZSC w/ Co-learn \cite{zhang+2024ijcv} & IJCV24 & FE / C-L14@336 & 98.9 & 93.2 & 81.0 & 83.0 & 98.6 & 98.8 & 95.7 & 84.8 & 84.8 & 97.3 & 95.1 & 41.6 & 88.6 \\
SHOT w/ Co-learn++ \cite{zhang+2024ijcv} & IJCV24 & ViL / C-L14@336 & 97.2 & 91.3 & 83.8 & 69.1 & 97.1 & 98.0 & 88.9 & 83.0 & 91.5 & 94.6 & 89.3 & 57.6 & 86.8 \\
SHOT++ w/ Co-learn++ \cite{zhang+2024ijcv} & IJCV24 & ViL / C-L14@336 & 97.4 & 89.4 & 88.0 & 86.0 & 98.0 & 96.4 & 93.9 & 85.2 & 97.8 & 94.5 & 94.3 & 45.0 & 88.8 \\
NRC w/ Co-learn++ \cite{zhang+2024ijcv} & IJCV24 & ViL / C-L14@336 & 98.0 & 90.8 & 83.9 & 69.0 & 97.4 & 97.6 & 91.7 & 81.6 & 92.8 & 96.2 & 92.8 & 59.9 & 87.6 \\
AaD w/ Co-learn++ \cite{zhang+2024ijcv} & IJCV24 & ViL / C-L14@336 & 97.7 & 92.1 & 87.1 & 83.5 & 98.1 & 98.3 & 93.7 & 85.8 & 95.4 & 95.6 & 94.0 & 64.0 & 90.4 \\
ZSC w/ Co-learn++ \cite{zhang+2024ijcv} & IJCV24 & ViL / C-L14@336 & 99.6 & \textbf{94.6} & 90.9 & 77.8 & \textbf{99.6} & 99.0 & 96.4 & 80.1 & 90.0 & 99.2 & 96.3 & 70.1 & 91.1 \\
DIFO \cite{tang+2024cvpr} & CVPR24 & ViL / C-B32 & 97.5 & 89.0 & 90.8 & 83.5 & 97.8 & 97.3 & 93.2 & 83.5 & 95.2 & 96.8 & 93.7 & 65.9 & 90.3 \\ 
LFTL \cite{lyu+2024eccv} & ECCV24 & MMA & 98.0 & 92.5 & 88.7 & \textbf{89.1} & 98.0 & 97.2 & 94.3 & \textbf{93.5} & \textbf{98.0} & 96.5 & 92.6 & 75.6 & \textbf{92.8} \\ 
\midrule
\rowcolor{red!15}
ViLAaD++ & - & ViL / C-B32 & 98.1 & 90.3 & 88.2 & 79.2 & 97.5 & 97.5 & 92.9 & 84.8 & 92.0 & 96.8 & 94.0 & 75.3 & 90.5 \\
\rowcolor{red!15}
ViLAaD++ & - & ViL / C-L14 & 98.5 & 94.0 & 91.1 & 78.9 & 98.5 & 98.6 & 95.3 & 86.8 & 93.9 & 97.8 & 95.6 & \textbf{76.4} & 92.1 \\
\rowcolor{red!15}
ViLAaD++ & - & ViL / C-L14@336 & 98.2 & 93.0 & 89.7 & 82.2 & 98.1 & \textbf{99.2} & 95.6 & 86.5 & 94.2 & 97.8 & 95.6 & 75.2 & 92.1 \\
\rowcolor{red!15}
ViLAaD++ & - & ViL / AL & 98.6 & 90.9 & 89.9 & 78.2 & 98.3 & 97.8 & 95.5 & 77.7 & 95.2 & 96.8 & 94.7 & 74.1 & 90.6 \\
\rowcolor{red!15}
ViLAaD++ & - & ViL / BL & 98.5 & 93.0 & 87.6 & 85.8 & 98.5 & 98.6 & 94.9 & 83.8 & 96.8 & 96.7 & 94.6 & 67.1 & 91.3     \\
\bottomrule
\end{tabular}
}
\caption{
C-SFDA results on VisDA-C \cite{peng+2017arxiv}. 
The table is divided into five blocks:
(1) Source model baseline,
(2) ZSC using ViL models,
(3) SFDA without auxiliary resources,
(4) SFDA leveraging auxiliary resources ({\it e.g.}, feature extractors (FE), minimal manual annotation (MMA), ViL models(ViL)),
(5) ViLAaD++ (proposed).
The best results are highlighted in bold.
}
\label{tab:eval:closed:visda}
\end{table*}
\begin{table*}[htbp]
\centering
\scalebox{0.9}{
\begin{tabular}{@{}l|l|l|cccccccccccc|c@{}}
\toprule
Method & Venue & Aux. & CP & CR & CS & PC & PR & PS & RC & RP & RS & SC & SP & SR & Avg. \\ \midrule
Source & - & - & 44.6 & 59.8 & 47.5 & 53.3 & 75.3 & 46.2 & 55.3 & 62.7 & 46.4 & 55.1 & 50.7 & 59.5 & 54.7 \\
\midrule
ZSC & - & ViL / C-B32 & 76.7 & 89.2 & 73.0 & 78.1 & 89.2 & 73.0 & 78.1 & 76.7 & 73.0 & 78.1 & 76.7 & 89.2 & 79.3 \\ 
ZSC & - & ViL / C-L14 & 87.3 & 93.8 & 86.6 & 87.8 & 93.8 & 86.6 & 87.8 & 87.3 & 86.6 & 87.8 & 87.3 & 93.8 & 88.9 \\ 
ZSC & - & ViL / C-L14@336 & 88.3 & 94.1 & 87.4 & 88.2 & 94.1 & 87.4 & 88.2 & 88.3 & 87.4 & 88.2 & 88.3 & 94.1 & 89.5 \\ 
ZSC & - & ViL / AL & 74.5 & 84.5 & 67.9 & 75.2 & 84.5 & 67.9 & 75.2 & 74.5 & 67.9 & 75.2 & 74.5 & 84.5 & 75.5 \\ 
ZSC & - & ViL / BL & 82.7 & 90.0 & 81.4 & 84.5 & 90.0 & 81.4 & 84.5 & 82.7 & 81.4 & 84.5 & 82.7 & 90.0 & 84.6 \\ 
\midrule
SHOT \cite{liang+2020icml} & ICML20 & - & 63.5 & 78.2 & 59.5 & 67.9 & 81.3 & 61.7 & 67.7 & 67.6 & 57.8 & 70.2 & 64.0 & 78.0 & 68.1     \\
GKD \cite{tang+2021iros} & IROS21 & - & 61.4 & 77.4 & 60.3 & 69.6 & 81.4 & 63.2 & 68.3 & 68.4 & 59.5 & 71.5 & 65.2 & 77.6 & 68.7     \\
NRC \cite{yang+2021neurips} & NeurIPS21 & - & 62.6 & 77.1 & 58.3 & 62.9 & 81.3 & 60.7 & 64.7 & 69.4 & 58.7 & 69.4 & 65.8 & 78.7 & 67.5     \\
AdaCon \cite{chen+2022cvpr} & CVPR22 & - & 60.8 & 74.8 & 55.9 & 62.2 & 78.3 & 58.2 & 63.1 & 68.1 & 55.6 & 67.1 & 66.0 & 75.4 & 65.4     \\
CoWA \cite{lee+2022icml} & ICML22 & - & 64.6 & 80.6 & 60.6 & 66.2 & 79.8 & 60.8 & 69.0 & 67.2 & 60.0 & 69.0 & 65.8 & 79.9 & 68.6     \\
PLUE \cite{litrico+2023cvpr} & CVPR23 & - & 59.8 & 74.0 & 56.0 & 61.6 & 78.5 & 57.9 & 61.6 & 65.9 & 53.8 & 67.5 & 64.3 & 76.0 & 64.7     \\
TPDS \cite{tang+2024ijcv} & IJCV24 & - & 62.9 & 77.1 & 59.8 & 65.6 & 79.0 & 61.5 & 66.4 & 67.0 & 58.2 & 68.6 & 64.3 & 75.3 & 67.1     \\
AaD w/DCPL \cite{diamant+2024eccv} & ECCV24 & - & 65.8 & 82.1 & 64.6 & 65.2 & 83.9 & 62.7 & 72.7 & 72.3 & 62.0 & 75.5 & 70.2 & 82.2 & 71.6 \\
SHOT w/ DCPL \cite{diamant+2024eccv} & ECCV24 & - & 62.9 & 80.6 & 61.7 & 63.3 & 82.1 & 60.7 & 70.1 & 71.0 & 59.8 & 73.4 & 68.7 & 80.7 & 69.6 \\
SHOT++ w/ DCPL \cite{diamant+2024eccv} & ECCV24 & - & 62.7 & 80.7 & 62.8 & 64.7 & 82.5 & 62.3 & 72.0 & 71.6 & 62.3 & 74.6 & 68.6 & 81.1 & 70.5 \\
\midrule
ZSC w/ Co-learn \cite{zhang+2024ijcv} & IJCV24 & FE / C-L14@336 & 75.1 & 86.5 & 78.5 & 78.9 & 86.7 & 76.8 & 85.4 & 79.1 & 76.7 & 81.2 & 73.8 & 84.4 & 80.3 \\
ZSC w/ Co-learn++ \cite{zhang+2024ijcv} & IJCV24 & ViL / C-L14@336 & 89.5 & 93.9 & \textbf{88.6} & \textbf{90.0} & 93.8 & \textbf{88.7} & 90.3 & 89.4 & \textbf{88.5} & 90.1 & 89.5 & 93.9 & 90.5 \\
DIFO \cite{tang+2024cvpr} & CVPR24 & ViL / C-B32 & 76.6 & 87.2 & 74.9 & 80.0 & 87.4 & 75.6 & 80.8 & 77.3 & 75.5 & 80.5 & 76.7 & 87.3 & 80.0     \\
\midrule
\rowcolor{red!15}
ViLAaD++ & - & ViL / C-B32 & 80.3 & 91.4 & 77.7 & 83.8 & 91.5 & 78.2 & 84.1 & 80.8 & 77.6 & 84.2 & 80.2 & 91.3 & 83.4 \\ 
\rowcolor{red!15}
ViLAaD++ & - & ViL / C-L14 & 89.1 & 95.1 & 87.8 & 89.9 & 95.1 & 87.8 & 89.5 & 89.1 & 87.9 & 90.1 & 89.2 & 95.1 & 90.5 \\ 
\rowcolor{red!15}
ViLAaD++ & - & ViL / C-L14@336 & \textbf{89.8} & \textbf{95.3} & 88.1 & \textbf{90.0} & \textbf{95.2} & 88.2 & \textbf{90.5} & \textbf{89.8} & 88.3 & \textbf{90.2} & \textbf{90.0} & \textbf{95.3} & \textbf{90.9} \\ 
\rowcolor{red!15}
ViLAaD++ & - & ViL / AL & 82.2 & 90.0 & 78.5 & 84.8 & 90.3 & 79.3 & 84.5 & 82.4 & 79.2 & 84.9 & 81.9 & 89.4 & 83.9 \\ 
\rowcolor{red!15}
ViLAaD++ & - & ViL / BL & 86.4 & 93.3 & 84.9 & 89.4 & 93.3 & 84.7 & 89.3 & 86.5 & 84.7 & 89.6 & 86.7 & 93.2 & 88.5 \\ 
\bottomrule
\end{tabular}
}
\caption{
C-SFDA results on DomainNet-126 \cite{peng+2019iccv,saito+2019iccv}. 
The table is divided into five blocks:
(1) Source model baseline,
(2) ZSC using ViL models,
(3) SFDA without auxiliary resources,
(4) SFDA leveraging auxiliary resources ({\it e.g.}, feature extractors (FE), ViL models (ViL)),
(5) ViLAaD++ (proposed).
The best results are highlighted in bold.
}
\label{tab:eval:closed:domainnet}
\end{table*}
\subsection{Evaluation Metrics.}
\label{sec:eval:metric}
For the C-SFDA and P-SFDA scenarios, following prior work, we report overall accuracy and its average across all domain shifts as evaluation metrics for the Office-31, Office-Home, and DomainNet-126 datasets.
For the VisDA-C dataset, we instead adopt per-class accuracy and report its mean.
In the O-SFDA setting, consistent with \cite{bucci+2020eccv,wan+2024cvpr,yang+2022neurips}, we use three metrics: overall accuracy on known classes (OS*), accuracy on the unknown class (UNK), and their harmonic mean (HOS), defined as
$\text{HOS} = \frac{2 \times \text{OS*} \times \text{UNK}}{\text{OS*} + \text{UNK}}$.
\subsection{Models.}
\label{sec:eval:model}
\noindent {\bf{Source / Target Models}}.
To ensure fair comparisons with prior works \cite{liang+2020icml, yang+2022neurips, tang+2024cvpr}, we adopt ResNet-50 \cite{he+2016cvpr} as the backbone architecture for both source and target models in experiments on the Office-31, Office-Home, and DomainNet-126 datasets. 
For the VisDA-C dataset, we use ResNet-101.
In the C-SFDA setting, we directly use the source model weights provided by DIFO\footnote{\url{https://github.com/tntek/source-free-domain-adaptation}} \cite{tang+2024cvpr}.
For the P-SFDA and O-SFDA scenarios, we follow the procedure of \cite{liang+2020icml} to train the source models, as pretrained weights for these settings are not publicly available.
The performance of our trained source models on each target domain is reported as ``Source*'' in Tables \ref{tab:eval:partial:oh} and \ref{tab:eval:open:oh}.
Notably, in the P-SFDA setting, our source models achieve an average accuracy that is 3.2\% lower than the source models used in prior work \cite{tang+2024cvpr} (denoted as ``Source'' in Table \ref{tab:eval:partial:oh}).
\par
\vspace{1mm}
\noindent {\bf{Vision-and-Language (ViL) Models}}.
In this work, we leverage five ViL models to construct the ViLAaD and ViLAaD++ instantiations: CLIP-ViT-B/32 (C-B32) \cite{radford+2021arxiv}, CLIP-ViT-L/14 (C-L14) \cite{radford+2021arxiv}, CLIP-ViT-L/14 with an input resolution of $336 \times 336$ (C-L14@336) \cite{radford+2021arxiv}, ALBEF (AL) \cite{li+2022icml}, and BLIP (BL) \cite{li+2022icml}.
All models are used via the LAVIS library\footnote{\url{https://github.com/salesforce/LAVIS}} \cite{li+2023acl}.
Although 
AL and BL
are capable of multimodal feature extraction from image-text pairs, we do not exploit this functionality in our approach.
All ViL models can be employed as zero-shot classifiers (ZSCs).
Their ZSC performance on each dataset is reported in the second block of 
Tables \ref{tab:eval:closed:o31}-\ref{tab:eval:closed:domainnet}, \ref{tab:eval:partial:oh} and \ref{tab:eval:open:oh}.
\subsection{Implementation Details}
\label{sec:eval:imple}
We train all models for 15 epochs using a batch size of 64 and the SGD optimizer with a momentum of 0.9 across all datasets.
For ZSC with ViL models, we adopt the commonly used text prompt format:
{\texttt{`a photo of a [CLS]'}},
where \texttt{[CLS]} denotes the class name.
All experiments are implemented in PyTorch and conducted on an NVIDIA A100 GPU.
\par
The main hyperparameters of ViLAaD and ViLAaD++ include $\lambda$ in Equation~\ref{eq:vilaad:loss}, the number of closed neighbors 
$K$, 
and the weights $\lambda^{\text{cls}}$ and $\lambda^{\text{div}}$ in Equation~\eqref{eq:vilaadpp:all}.
Following AaD~\cite{yang+2022neurips}, we decay $\lambda$ during adaptation according to 
$\lambda = 1 + (10 \times \frac{\text{iter}}{\text{max\_iter}})^{-\beta}$,
where $\beta$ controls the rate of decay. 
Based on grid search results, we set $\beta = 1.0$ for Office-31, Office-Home and DomainNet-126, and $\beta = 4.0$ for VisDA-C.
For the number of closed neighbors $K$, we use $K=3$ for Office-31, $K=2$ for Office-Home and DomainNet-126, and $K=5$ for VisDA-C. 
We adopt $\lambda^{\text{cls}} = 0.5$ and $\lambda^{\text{div}} = 1.0$ as default hyperparameter settings throughout the paper.
However, in the P-SFDA scenario, we set $\lambda^{\text{div}} = 0.0$, since the target domain contains only a subset of the classes present in the source domain \cite{liang+2020icml}.
Detailed analyses of the hyperparameters $K$, $\lambda^{\text{cls}}$, and $\lambda^{\text{div}}$ are presented in \S\ref{sec:eval:analysis}.
\subsection{Evaluation of ViLAaD}
\label{sec:eval:vilaad}
Table \ref{tab:eval:vilaad:c-b32} (a)-(c) presents the classification performance of the source model without adaptation (denoted as ``Source''), zero-shot classification with a ViL model (denoted as ``ZSC''), AaD \cite{yang+2022neurips}, and our proposed ViLAaD. 
The evaluations are conducted on the Office-31 \cite{saenko+eccv2010}, Office-Home \cite{venkateswara+2017cvpr}, and VisDA-C \cite{peng+2017arxiv} datasets. 
Both ZSC and ViLAaD utilize C-B32 to generate the results. 
These tables demonstrate that ViLAaD consistently outperforms other methods in nearly all cases.
An exception occurs in the DW scenario ({\it i.e.}, the source domain is \uline{D}SLR and the target domain is \uline{W}eb) on the Office-31 dataset, where the ZSC performance is significantly worse than that of the Source. 
In this case, the ViL model does not serve as a ``reasonable initialization'' for the target dataset, resulting in lower performance than AaD \cite{yang+2022neurips}, which does not rely on the ViL model for adaptation.
\par
Figure \ref{fig:eval:vilaad-vils} presents classification results using four different ViL models: C-B32, C-L14, AL, and BL.
For improved interpretability, we report only the average accuracy across all source-target domain pairs for the Office-31 and Office-Home datasets.
The results show that ViLAaD consistently outperforms Source.
Furthermore, ViLAaD surpasses Source, ZSC, and AaD in 6 out of 12 settings, despite using significantly smaller adaptation models (ResNet-50 or ResNet-101) compared to C-B32.
While it does not achieve the highest performance in the remaining cases, these instances can be further categorized as follows:
\par
\vspace{1mm}
\noindent \textbf{Pattern I: ViLAaD outperforms Source and AaD but falls short of ZSC.}
This pattern is observed on the Office-Home dataset when using C-L14 and BL.
The key contributing factor is the substantial performance gap between Source and ZSC: 
 ZSC with C-L14 achieves 27.5\% higher accuracy than Source while ZSC with BL shows a 25.4\% gain on average.
In such cases, the source model provides a weak initialization for ZSC, limiting the synergy between the source and ViL models. 
As a result, ViLAaD struggles to surpass the strong standalone performance of ZSC.
\par
\vspace{1mm}
\noindent \textbf{Pattern II: ViLAaD outperforms Source and ZSC but lags behind AaD.}
This pattern is observed on the Office-31 and Office-Home datasets when using AL, as well as on the VisDA-C dataset when using AL and BL.
In the first two cases, the corresponding ViL model, AL, does not provide a strong initialization: as shown in Figure~\ref{fig:eval:vilaad-vils}, ZSC yields performance that is lower than or comparable to the source model.
However, in the remaining cases, ZSC demonstrates much better performance than Source.
To further explore this behavior, we analyze the per-class accuracies of the Source, ZSC, and ViLAaD models.
Figure~\ref{fig:eval:conf_mat} presents confusion matrices for Source, ZSC using AL, and ViLAaD with AL, respectively.
They reveal that for certain classes such as ``person'' (per) and ``car'', ZSC performs notably worse than Source ({\it e.g.}, 13.4\% vs. 20.0\% for the ``person'' class, and 55.0\% vs. 67.9\% for the ``car'' class).
Since ViLAaD does not include mechanisms to improve the ZSC baseline itself, its overall performance can be negatively impacted when the ViL model struggles with specific classes (see Figure~\ref{fig:eval:conf_mat}~(c)).
This can lead to ViLAaD performing worse than AaD, which, despite not using ViL models, incorporates mechanisms to directly enhance the predictions of the source model.
\par
\vspace{1mm}
However, as shown in Figure~\ref{fig:eval:vilaad-vils}, the enhanced variant ViLAaD++ consistently outperforms ViLAaD and achieves the best performance among all baselines in 11 out of 12 cases.
This improvement can be attributed to: (1) the additional objectives for target model adaptation in ViLAaD++, which better leverage the synergy between the source and ViL models, and (2) prompt tuning of the ViL model, which enhances ZSC performance and mitigates the limitations of ViLAaD observed in Pattern II.
In the next subsection, we compare ViLAaD++ with existing state-of-the-art methods.
\subsection{Evaluation of ViLAaD++}
\label{sec:eval:vilaadpp}
Tables~\ref{tab:eval:closed:o31}, \ref{tab:eval:closed:oh}, \ref{tab:eval:closed:visda}, and \ref{tab:eval:closed:domainnet} present the performance of state-of-the-art methods under the C-SFDA scenario on the Office-31, Office-Home, VisDA-C, and DomainNet-126 datasets, respectively.
Each table is organized into five blocks:
\begin{enumerate}
\item Source Model: Reports the baseline performance of the source-trained model.
\item ZSC: Shows zero-shot classification results obtained using ViL models.
\item SFDA without Auxiliary Resources: Includes state-of-the-art SFDA methods that do not rely on any additional resources.
\item SFDA with Auxiliary Resources: Presents SFDA results that incorporate auxiliary inputs such as feature extractors (FE), minimal manual annotation (MMA), user feedback (UF), and ViL models (ViL).
\item ViLAaD++: Displays the performance of our proposed ViLAaD++ method using C-B32, C-L14, C-L14@336, AL, and BL.
\end{enumerate}
Notice that results in the second block are directly obtained from the respective ViL models, while those in all other blocks are based on inference using ResNet-50 \cite{he+2016cvpr}.
\par
Compared to DIFO \cite{tang+2024cvpr}, which also employs C-B32, ViLAaD++ with C-B32 consistently outperforms it across all datasets.
Specifically, ViLAaD++ achieves gains of 0.1\%, 0.8\%, and 3.4\% in average accuracy on the Office-31, Office-Home, and DomainNet-126 datasets, respectively, and improves mean per-class accuracy on VisDA-C by 0.2\%.
Among all ViLAaD++ variants, the configuration using C-L14@336 delivers the best performance across all datasets.
It achieves the highest average accuracy on Office-31 (0.1\% higher than the previous best, LFTL \cite{lyu+2024eccv}), Office-Home (3.5\% higher than ZSC w/ Co-learn++ \cite{zhang+2024ijcv}), and DomainNet-126 (0.4\% higher than ZSC w/ Co-learn++ \cite{zhang+2024ijcv}).
While the best-performing ViLAaD++ variants (using C-L14 or C-L14@336) slightly trail the top-performing method, LFTL, in mean per-class accuracy on the VisDA-C dataset, they still achieve higher per-class accuracy in 9 out of 12 categories with C-L14 and in 8 out of 12 categories with C-L14@336, respectively.
Overall, these results demonstrate that ViLAaD++ effectively enhances cross-domain performance in the C-SFDA scenario, outperforming or matching the state-of-the-art across a wide range of benchmarks.
\begin{table}[]
\centering
\begin{subtable}{0.22\textwidth}
\centering
\scalebox{1.0}{
\begin{tabular}{l|cc}
\toprule
$K$ & ViLAaD & ViLAaD++ \\ \midrule
1 & 77.9 & 83.6 \\
\cellcolor{red!15}2 & \cellcolor{red!15}\textbf{80.0} & \cellcolor{red!15}\textbf{83.9} \\
3 & 79.3 & 83.9 \\
4 & 78.5 & 83.8 \\
5 & 77.7 & 83.7 \\
\bottomrule
\end{tabular}
}
\caption{}
\end{subtable}
\begin{subtable}{0.25\textwidth}
\centering
\scalebox{1.0}{
\begin{tabular}{ll|ccc}
\toprule
                         &                          & \multicolumn{3}{c}{$\lambda^{\text{cls}}$} \\
                         &                          & 0.4     & \cellcolor{red!15}0.5     & 0.6    \\ \midrule 
\multirow{3}{*}{$\lambda^{\text{div}}$} & \multicolumn{1}{l|}{0.9} & 83.7    & 83.7    & 83.8     \\
                         & \multicolumn{1}{l|}{\cellcolor{red!15}1.0} & 83.7    & \cellcolor{red!15}\textbf{83.9}    & 83.8     \\
                         & \multicolumn{1}{l|}{1.1} & 83.8    & 83.6    & 83.8 \\ \bottomrule  
\end{tabular}
}
\caption{}
\end{subtable}
\caption{Parameter studies for (a) the number of nearest neighbors $K$ and (b) the loss weights $\lambda_{\text{cls}}$ and $\lambda_{\text{div}}$. All scores are the average accuracies on the Office-Home dataset.}
\label{tab:eval:param}
\end{table}
\subsection{Model Analysis}
\label{sec:eval:analysis}
Tables~\ref{tab:eval:param} (a) and (b) present the average accuracies of ViLAaD and ViLAaD++ on the Office-Home dataset under different parameter settings.
In Table~\ref{tab:eval:param} (a), we evaluate the impact of the number of neighbors $K$ used to define the closed neighbor set $\mathcal{C}$.
The results show that performance is generally stable across different values of $K$, with the best results consistently achieved at $K=2$ for both ViLAaD and ViLAaD++.
Table~\ref{tab:eval:param} (b) examines the effect of the auxiliary loss weights $\lambda^{\text{cls}}$ and $\lambda^{\text{div}}$ introduced in the ViLAaD++ target model adaptation (see Equation~\eqref{eq:vilaadpp:all}).
Again, the results indicate that varying these parameters does not lead to significant changes in performance.
For the experiments in \S\ref{sec:eval:vilaadpp}, we use $\lambda^{\text{cls}} = 0.5$ and $\lambda^{\text{div}} = 1.0$, as this setting yields the best observed performance.
\par
Table~\ref{tab:eval:ablation} presents ablation results evaluating the contributions of the individual loss components introduced in ViLAaD++: $L^{\text{pro}}$ for ViL prompt tuning, and $L^{\text{cls}}$ and $L^{\text{div}}$ for target model adaptation.
The results show that each loss component contributes complementarily to improving SFDA performance.
The highest average accuracy is achieved when all loss terms are used, which corresponds to the full ViLAaD++ configuration.
\par
\begin{figure}[t]
\centering
\includegraphics[width=0.5\textwidth, page=1]{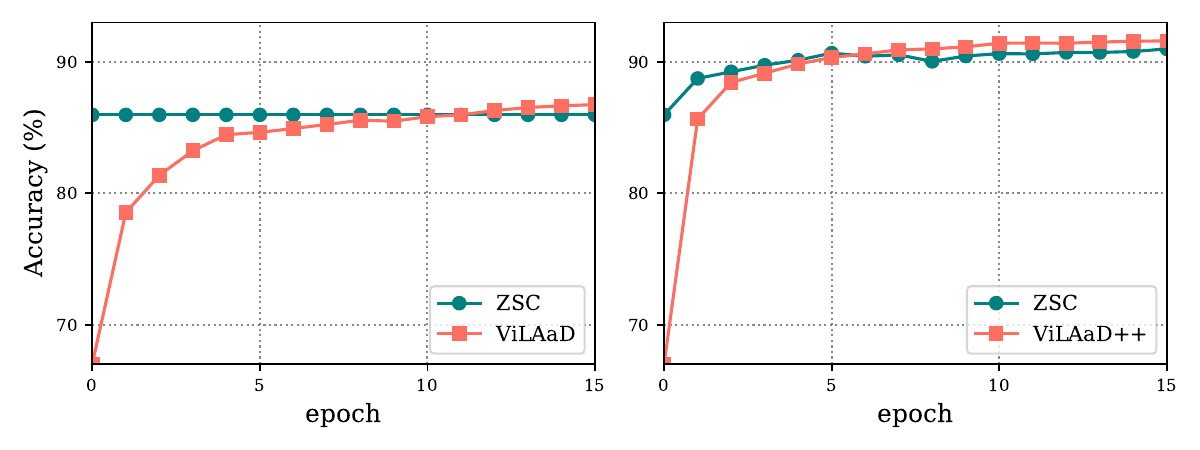}
\caption{
Accuracies of ViLAaD (left) and ViLAaD++ (right) with respect to the number of epochs in the AP scenario on the Office-Home dataset.
C-B32 is used to generate the results.
Note that in ViLAaD, the accuracy of zero-shot classification (ZSC) by the ViL model remains unchanged throughout training, as both the ViL model and its text prompt are kept frozen.
In contrast, ViLAaD++ shows improved ZSC accuracy over time, as its text prompt is jointly tuned during adaptation.
}
\label{fig:eval:epoch}
\end{figure}

Figure~\ref{fig:eval:epoch} shows the accuracies of ViLAaD and ViLAaD++ over the course of training epochs.
In both cases, accuracy improves steadily and eventually surpasses the zero-shot classification (ZSC) performance of the ViL model (C-B32 in this case).
Notably, in ViLAaD++, the ZSC accuracy itself also increases over time, as the text prompt is jointly tuned during adaptation.
Interestingly, the adapted model in ViLAaD++ surpasses the ZSC baseline earlier than in ViLAaD, despite the ZSC performance also improves.
These results demonstrate that both ViLAaD and ViLAaD++ effectively adapt to the target dataset through optimization, with ViLAaD++ benefiting additionally from prompt tuning.
\par
\begin{figure*}[t]
\centering
\includegraphics[width=1.0\textwidth, page=1]{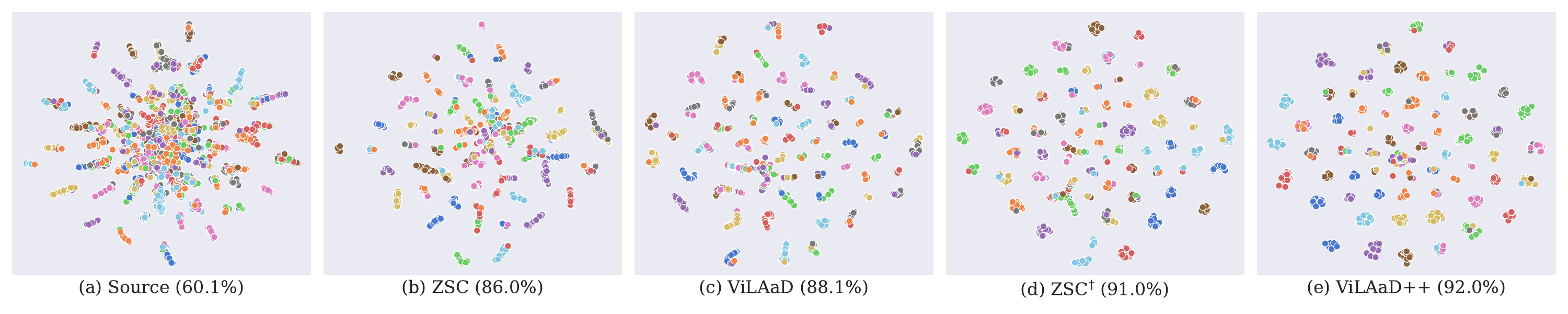}
\caption{
t-SNE visualizations of the predicted probability distributions ({\it i.e.}, $p$ or $q$) for (a) source model classification, (b) zero-shot classification (ZSC) using a ViL model without prompt tuning, (c) ViLAaD, (d) ZSC using a ViL model with prompt tuning, and (e) ViLAaD++ in the CP scenario on the Office-Home dataset.
In all cases, C-B32 is used as the underlying ViL model.
Different colors correspond to different classes.
}
\label{fig:eval:tsne}
\end{figure*}

Figure~\ref{fig:eval:tsne} (a)–(e) show t-SNE visualizations of the predicted probability distributions ({\it i.e.}, $p$ from the adaptation model or $q$ from the ViL model) across different methods. 
As expected, comparing (a) and (c) reveals that optimizing the ViLAaD objective encourages examples from the same class to form more distinct clusters. 
This clustering effect becomes even more pronounced with ViLAaD++ optimization, as shown in (e).
Notably, (d) presents the ZSC output from a ViL model whose text prompts have been optimized via the ViLAaD++ framework (see \S\ref{sec:proposed:vilaadpp}). 
Compared to (b), which illustrates ZSC without prompt tuning, the tuned version exhibits more coherent class-wise clustering—closely resembling the behavior seen in ViLAaD and ViLAaD++.
These results underscore the importance of jointly optimizing the ViL text prompts along with the adaptation model.

\begin{table*}[]
\centering
\scalebox{1.0}{
\begin{tabular}{@{}l|ccc|cccccccccccc|l@{}}
\toprule
Method & $L^{\text{Pro}}$ & $L^{\text{Div}}$ & $L^{\text{Cls}}$ & AC & AP & AR & CA & CP & CR & PA & PC & PR & RA & RC & RP & Avg. \\ \midrule
ViLAaD &  &  &  & 66.4 & 86.8 & 87.5 & 77.2 & 88.1 & 87.1 & 77.1 & 65.9 & 88.5 & 78.7 & 68.0 & 89.1 & 80.0     \\
               & \checkmark &  &  & 68.2 & 88.7 & 88.3 & 78.2 & 89.9 & 87.6 & 77.3 & 68.3 & 88.9 & 80.3 & 70.2 & 91.3 & 81.4    \\
               & \checkmark & \checkmark &  & 69.3 & 90.6 & 88.5 & 79.2 & 90.3 & 88.0 & 78.2 & 69.9 & 88.8 & 79.6 & 70.3 & 91.1 & 82.0    \\
               & \checkmark & & \checkmark & 69.6 & \textbf{91.6} & \textbf{90.0} & 82.9 & 91.5 & 89.8 & 80.7 & 69.8 & \textbf{90.1} & 82.9 & \textbf{71.4} & 92.0 & 83.5    \\
ViLAaD++ & \checkmark & \checkmark & \checkmark & \textbf{70.1} & \textbf{91.6} & 89.9 & \textbf{83.2} & \textbf{92.0} & \textbf{90.0} & \textbf{81.0} & \textbf{71.7} & 89.9 & \textbf{83.3} & 71.3 & \textbf{92.3} & \textbf{83.9} \\
\bottomrule
\end{tabular}
}
\caption{Ablation studies for the auxiliary losses composing ViLAaD++: $L^{\text{pro}}$, $L^{\text{cls}}$ and $L^{\text{div}}$. All scores are obtained by the experiment on the Office-Home dataset.}
\label{tab:eval:ablation}
\end{table*}
\subsection{Evaluation in P-SFDA \& O-SFDA Scenarios}
\label{sec:eval:po}
Table \ref{tab:eval:partial:oh} presents the Partial-set SFDA (P-SFDA) results of state-of-the-art methods on the Office-Home dataset. 
Similar to the C-SFDA scenario, we report the performance of ViLAaD++ using C-B32, C-L14, C-L14@336, AL and BL.
As noted in \S\ref{sec:eval:model}, the source models used to initialize ViLAaD++ (denoted as ``Source*'' in Table \ref{tab:eval:partial:oh}) generally perform worse than those employed by existing methods (denoted as ``Source''). Despite this, ViLAaD++ with C-B32 outperforms the average accuracy of DIFO \cite{tang+2024cvpr}, which also uses C-B32, by 2.6 \%.
Consistent with the C-SFDA results, ViLAaD++ achieves its best performance with C-L14@336, outperforming the previous best ({\it i.e.}, DIFO) by a significant margin of 7.8\% in average accuracy. 
Interestingly, even ZSC using C-L14 or C-L14@336 surpasses existing methods (85.9\% or 87.1\% vs. 85.6\% average accuracy). 
Nevertheless, ViLAaD++ consistently outperforms the corresponding ZSC results, despite relying on a much smaller ResNet50-based classifier compared to ViL models.
These findings highlight the strong effectiveness of ViLAaD++ in the P-SFDA setting.
\par
Table~\ref{tab:eval:open:oh} presents the Open-set SFDA (O-SFDA) results of state-of-the-art methods on the Office-Home dataset.
We also report the performance of ViLAaD++ using the same ViL models as in the C-SFDA and P-SFDA scenarios.
Compared to the previous best method, UPUK \cite{wan+2024cvpr}, ViLAaD++ with C-L14@336 achieves a 4.7-point higher average HOS score.
However, except the case with AL, the HOS scores of ViLAaD++ fall behind the corresponding ZSC results.
Although the adaptation model (ResNet50) is significantly smaller than the ViL models, a possible reason for this performance gap is the large disparity between the source model and the ViL model, similar to the discussion in \S\ref{sec:eval:vilaad}.
For instance, as shown in Table~\ref{tab:eval:open:oh}, the source model achieves only 55.3 HOS, whereas ZSC with C-L14@336 reaches 80.6.
\begin{table*}[htbp]
\centering
\scalebox{0.9}{
\begin{tabular}{@{}l|l|l|cccccccccccc|c@{}}
\toprule
Method & Venue & Aux. & AC & AP & AR & CA & CP & CR & PA & PC & PR & RA & RC & RP & Avg. \\ \midrule
Source \cite{tang+2024cvpr} & - & - & 45.2 & 70.4 & 81.0 & 56.2 & 60.8 & 66.2 & 60.9 & 40.1 & 76.2 & 70.8 & 48.5 & 77.3 & 62.8 \\ 
Source* & - & - & 41.0 & 64.4 & 72.1 & 57.0 & 62.6 & 64.8 & 58.0 & 37.5 & 72.5 & 66.6 & 45.5 & 73.5 & 59.6 \\ 
\midrule
ZSC & - & ViL / C-B32 & 59.3 & 85.0 & 87.3 & 81.1 & 85.0 & 87.3 & 81.1 & 59.3 & 87.3 & 81.1 & 59.3 & 85.0 & 78.2 \\
ZSC & - & ViL / C-L14 & 72.5 & 88.7 & 92.8 & 89.6 & 88.7 & 92.8 & 89.6 & 72.5 & 92.8 & 89.6 & 72.5 & 88.7 & 85.9 \\
ZSC & - & ViL / C-L14@336 & 74.4 & 90.1 & 93.6 & 90.6 & 90.1 & 93.6 & 90.6 & 74.4 & 93.6 & 90.6 & 74.4 & 90.1 & 87.1 \\
ZSC & - & ViL / AL & 44.4 & 64.9 & 64.5 & 58.2 & 64.9 & 64.5 & 58.2 & 44.4 & 64.5 & 58.2 & 44.4 & 64.9 & 58.0 \\ 
ZSC & - & ViL / BL & 73.5 & 88.6 & 86.5 & 86.1 & 88.6 & 86.5 & 86.1 & 73.5 & 86.5 & 86.1 & 73.5 & 88.6 & 83.7 \\ 
\midrule
SHOT \cite{liang+2020icml} & ICML20 & - & 64.8 & 85.2 & 92.7 & 76.3 & 77.6 & 88.8 & 79.7 & 64.3 & 89.5 & 80.6 & 66.4 & 85.8 & 79.3 \\
HCL \cite{huang+2022neurips} & NeurIPS21 & - & 65.6 & 85.2 & 92.7 & 77.3 & 76.2 & 87.2 & 78.2 & 66.0 & 89.1 & 81.5 & 68.4 & 87.3 & 79.6 \\
CoWA \cite{lee+2022icml} & ICML22 & - & 69.6 & 93.2 & 92.3 & 78.9 & 81.3 & 92.1 & 79.8 & 71.7 & 90.0 & 83.8 & 72.2 & 93.7 & 83.2 \\
AaD \cite{yang+2022neurips} & NIPS22 & - & 67.0 & 83.5 & 93.1 & 80.5 & 76.0 & 87.6 & 78.1 & 65.6 & 90.2 & 83.5 & 64.3 & 87.3 & 79.7 \\
CRS \cite{zhang+2023cvpr} & CVPR23 & - & 68.6 & 85.1 & 90.9 & 80.1 & 79.4 & 86.3 & 79.2 & 66.1 & 90.5 & 82.2 & 69.5 & 89.3 & 80.6 \\
\midrule
SHOT w/ Co-learn \cite{zhang+2024ijcv} & IJCV24 & FE / C-L14@336 & 67.0 & 87.3 & 91.8 & 77.0 & 71.2 & 85.2 & 77.3 & 70.2 & 91.3 & 84.0 & 71.3 & 87.8 & 80.1 \\
SHOT w/ Co-learn++ \cite{zhang+2024ijcv} & IJCV24 & ViL / C-L14@336 & 68.0 & 86.6 & 91.7 & 77.7 & 73.1 & 89.3 & 79.2 & 68.3 & 91.2 & 83.3 & 68.2 & 87.5 & 80.3 \\
DIFO \cite{tang+2024cvpr} & CVPR24 & ViL / C-B32 & 70.2 & 91.7 & 91.5 & 87.8 & 92.6 & 92.9 & 87.3 & 70.7 & 92.9 & 88.5 & 69.6 & 91.5 & 85.6 \\ 
\midrule
\rowcolor{red!15}
ViLAaD++ & - & ViL / C-B32 & 75.7 & 93.9 & 93.7 & 88.9 & 92.7 & 93.2 & 88.8 & 76.9 & 94.1 & 90.9 & 76.8 & 93.3 & 88.2 \\
\rowcolor{red!15}
ViLAaD++ & - & ViL / C-L14 & 85.8 & 96.5 & 96.6 & 94.0 & \textbf{96.7} & 96.3 & 93.5 & 85.8 & 96.4 & 94.0 & 85.1 & 96.4 & 93.1 \\
\rowcolor{red!15}
ViLAaD++ & - & ViL / C-L14@336 & 85.5 & 96.5 & \textbf{96.7} & \textbf{94.1} & 96.3 & \textbf{96.6} & \textbf{94.2} & \textbf{86.4} & \textbf{96.5} & \textbf{94.6} & \textbf{86.5} & \textbf{96.9} & \textbf{93.4} \\
\rowcolor{red!15}
ViLAaD++ & - & ViL / AL & 75.6 & 88.9 & 89.1 & 84.0 & 89.0 & 89.2 & 83.8 & 76.4 & 89.6 & 84.1 & 76.7 & 88.4 & 84.6 \\
\rowcolor{red!15}
ViLAaD++ & - & ViL / BL & \textbf{85.9} & \textbf{96.6} & 94.8 & 92.2 & 96.0 & 94.8 & 92.7 & 85.9 & 94.8 & 93.5 & 86.1 & 96.5 & 92.5 \\
\bottomrule
\end{tabular}
}
\caption{
Partial-set SFDA (P-SFDA) results on Office-Home.
The table is divided into five blocks:
(1) Source model baseline,
(2) ZSC using ViL models,
(3) SFDA without auxiliary resources,
(4) SFDA leveraging auxiliary resources ({\it e.g.}, feature extractors (FE), ViL models (ViL)),
(5) ViLAaD++ (proposed).
The best results are highlighted in bold.
}
\label{tab:eval:partial:oh}
\end{table*}
\begin{table*}[htbp]
\centering
\scalebox{0.7}{
\begin{tabular}{@{}l|l|ccc|ccc|ccc|ccc|ccc|ccc|ccc@{}}
\toprule
Method & Aux. & \multicolumn{3}{c|}{AC} & \multicolumn{3}{c|}{AP} & \multicolumn{3}{c|}{AR} & \multicolumn{3}{c|}{CA} & \multicolumn{3}{c|}{CP}& \multicolumn{3}{c|}{CR} & & & \\
       &    & OS* & UNK & HOS & OS* & UNK & HOS & OS* & UNK & HOS & OS* & UNK & HOS & OS* & UNK & HOS & OS* & UNK & HOS &  &  &  \\
\midrule
Source* & - & 30.4 & 87.6 & 45.2 & 48.3 & 87.3 & 62.2 & 57.4 & 87.3 & 69.3 & 35.9 & 83.8 & 50.3 & 40.8 & 84.2 & 55.0 & 48.6 & 85.0 & 61.8 &  &  &  \\
\midrule
ZSC & ViL / C-B32 & 56.7 & 75.7 & 64.8 & 83.7 & 75.9 & 79.6 & 86.7 & 77.5 & 81.8 & 73.5 & 80.6 & 76.9 & 83.7 & 75.9 & 79.6 & 86.7 & 77.5 & 81.8 &  &  & \\
ZSC & ViL / C-L14 & 73.2 & 72.8 & 73.0 & 91.8 & 70.5 & 79.8 & 95.3 & 72.8 & 82.6 & 86.2 & 79.7 & 82.8 & 91.8 & 70.5 & 79.8 & 95.3 & 72.8 & 82.6 &  &  &  \\
ZSC & ViL / C-L14@336 & 74.3 & 74.1 & 74.2 & \textbf{92.5} & 71.4 & 80.6 & \textbf{95.8} & 73.5 & \textbf{83.7} & \textbf{87.6} & 80.5 & \textbf{83.9} & \textbf{92.5} & 71.4 & 80.6 & \textbf{95.8} & 73.5 & \textbf{83.7} &  &  &  \\
ZSC & ViL / AL & 28.9 & \textbf{87.1} & 43.4 & 51.5 & \textbf{92.7} & 66.2 & 48.2 & \textbf{93.2} & 63.6 & 32.6 & \textbf{92.1} & 48.2 & 51.5 & \textbf{92.7} & 66.2 & 48.2 & \textbf{93.2} & 63.6 &  &  &  \\
ZSC & ViL / BL & \textbf{75.3} & 76.7 & \textbf{76.0} & 89.2 & 78.0 & \textbf{83.2} & 85.9 & 79.7 & 82.7 & 81.0 & 76.6 & 78.7 & 89.2 & 78.0 & \textbf{83.2} & 85.9 & 79.7 & 82.7 &  &  &  \\
\midrule
Cluster      & - & 6.0 & 63.0 & 11.0 & 5.0 & 55.0 & 10.0 & 4.0 & 62.0 & 8.0 & 8.0 & 59.0 & 14.0 & 5.0 & 55.0 & 9.0 & 5.0 & 56.0 & 8.0 &  &  &  \\
SHOT \cite{liang+2020icml} & - & 67.0 & 28.0 & 39.5 & 81.8 & 26.3 & 39.8 & 87.5 & 32.1 & 47.0 & 66.8 & 46.2 & 54.6 & 77.5 & 27.2 & 40.2 & 80.0 & 25.9 & 39.1 &  &  &  \\
AaD \cite{yang+2022neurips} & - & 50.5 & 67.4 & 57.7 & 64.0 & 66.4 & 65.1 & 72.2 & 69.5 & 70.8 & 47.1 & 80.3 & 59.3 & 64.7 & 68.2 & 66.4 & 65.0 & 71.0 & 67.8 & &  &  \\
UPUK \cite{wan+2024cvpr} & - & 49.0 & 64.8 & 55.8 & 68.1 & 88.0 & 76.7 & 71.5 & 86.6 & 78.4 & 61.3 & 72.6 & 66.4 & 66.5 & 81.6 & 73.1 & 71.6 & 84.8 & 77.6 &  &  &  \\
\midrule
\cellcolor{red!15}ViLAaD++ & \cellcolor{red!15}ViL / C-B32 & \cellcolor{red!15}61.1 & \cellcolor{red!15}63.0 & \cellcolor{red!15}62.0 & \cellcolor{red!15}78.5 & \cellcolor{red!15}68.8 & \cellcolor{red!15}73.3 & \cellcolor{red!15}81.4 & \cellcolor{red!15}64.9 & \cellcolor{red!15}72.2 & \cellcolor{red!15}65.4 & \cellcolor{red!15}64.4 & \cellcolor{red!15}64.9 & \cellcolor{red!15}71.7 & \cellcolor{red!15}68.2 & \cellcolor{red!15}69.9 & \cellcolor{red!15}74.0 & \cellcolor{red!15}67.1 & \cellcolor{red!15}70.4 &  &  &  \\
\cellcolor{red!15}ViLAaD++ & \cellcolor{red!15}ViL / C-L14 & \cellcolor{red!15}66.9 & \cellcolor{red!15}72.6 & \cellcolor{red!15}69.6 & \cellcolor{red!15}79.7 & \cellcolor{red!15}74.0 & \cellcolor{red!15}76.8 & \cellcolor{red!15}83.6 & \cellcolor{red!15}73.1 & \cellcolor{red!15}78.0 & \cellcolor{red!15}72.1 & \cellcolor{red!15}74.7 & \cellcolor{red!15}73.4 & \cellcolor{red!15}78.0 & \cellcolor{red!15}70.2 & \cellcolor{red!15}73.9 & \cellcolor{red!15}81.1 & \cellcolor{red!15}65.6 & \cellcolor{red!15}72.5 &  &  &  \\
\cellcolor{red!15}ViLAaD++ & \cellcolor{red!15}ViL / C-L14@336 & \cellcolor{red!15}69.6 & \cellcolor{red!15}71.9 & \cellcolor{red!15}70.7 & \cellcolor{red!15}83.2 & \cellcolor{red!15}74.6 & \cellcolor{red!15}78.6 & \cellcolor{red!15}89.1 & \cellcolor{red!15}65.5 & \cellcolor{red!15}75.5 & \cellcolor{red!15}72.2 & \cellcolor{red!15}73.1 & \cellcolor{red!15}72.7 & \cellcolor{red!15}79.7 & \cellcolor{red!15}68.1 & \cellcolor{red!15}73.4 & \cellcolor{red!15}75.6 & \cellcolor{red!15}72.8 & \cellcolor{red!15}74.2 &  &  &  \\
\cellcolor{red!15}ViLAaD++ & \cellcolor{red!15}ViL / AL & \cellcolor{red!15}53.1 & \cellcolor{red!15}70.6 & \cellcolor{red!15}60.6 & \cellcolor{red!15}71.4 & \cellcolor{red!15}80.1 & \cellcolor{red!15}75.5 & \cellcolor{red!15}74.1 & \cellcolor{red!15}72.2 & \cellcolor{red!15}73.1 & \cellcolor{red!15}64.3 & \cellcolor{red!15}75.8 & \cellcolor{red!15}69.6 & \cellcolor{red!15}75.6 & \cellcolor{red!15}75.9 & \cellcolor{red!15}75.7 & \cellcolor{red!15}78.5 & \cellcolor{red!15}61.9 & \cellcolor{red!15}69.2 &  &  &  \\
\cellcolor{red!15}ViLAaD++ & \cellcolor{red!15}ViL / BL & \cellcolor{red!15}69.9 & \cellcolor{red!15}66.8 & \cellcolor{red!15}68.3 & \cellcolor{red!15}90.2 & \cellcolor{red!15}72.5 & \cellcolor{red!15}80.4 & \cellcolor{red!15}78.4 & \cellcolor{red!15}71.4 & \cellcolor{red!15}74.7 & \cellcolor{red!15}65.8 & \cellcolor{red!15}78.5 & \cellcolor{red!15}71.6 & \cellcolor{red!15}79.7 & \cellcolor{red!15}69.8 & \cellcolor{red!15}74.4 & \cellcolor{red!15}74.9 & \cellcolor{red!15}69.4 & \cellcolor{red!15}72.1 &  &  &  \\
\bottomrule
Method & Aux. & \multicolumn{3}{c|}{PA} & \multicolumn{3}{c|}{PC} & \multicolumn{3}{c|}{PR} & \multicolumn{3}{c|}{RA} & \multicolumn{3}{c|}{RC} & \multicolumn{3}{c|}{RP} & \multicolumn{3}{c}{Avg.} \\
       &    & OS* & UNK & HOS & OS* & UNK & HOS & OS* & UNK & HOS & OS* & UNK & HOS & OS* & UNK & HOS & OS* & UNK & HOS & OS* & UNK & HOS \\
\midrule
Source*      & - & 34.5 & 87.3 & 49.5 & 21.7 & 85.8 & 34.6 & 54.5 & 87.6 & 67.2 & 45.8 & 82.8 & 59.0 & 28.0 & 81.3 & 41.7 & 57.3 & 82.9 & 67.8 & 41.9 & 85.2 & 55.3 \\
\midrule
ZSC & ViL / C-B32 & 73.5 & 80.6 & 76.9 & 56.7 & 75.7 & 64.8 & 86.7 & 77.5 & 81.8 & 73.5 & 80.6 & 76.9 & 56.7 & 75.7 & 64.8 & 83.7 & 75.9 & 79.6 & 75.1 & 77.4 & 75.8 \\
ZSC & ViL / C-L14 & 86.2 & 79.7 & 82.8 & 73.2 & 72.8 & 73.0 & 95.3 & 72.8 & 82.6 & 86.2 & 79.7 & 82.8 & 73.2 & 72.8 & 73.0 & 91.8 & 70.5 & 79.8 & 86.6 & 74.0 & 79.5 \\
ZSC & ViL / C-L14@336 & \textbf{87.6} & 80.5 & \textbf{83.9} & 74.3 & 74.1 & 74.2 & \textbf{95.8} & 73.5 & \textbf{83.7} & \textbf{87.6} & 80.5 & \textbf{83.9} & 74.3 & 74.1 & 74.2 & \textbf{92.5} & 71.4 & 80.6 & \textbf{87.6} & 74.9 & \textbf{80.6} \\
ZSC & ViL / AL & 32.6 & \textbf{92.1} & 48.2 & 28.9 & \textbf{87.1} & 43.4 & 48.2 & \textbf{93.2} & 63.6 & 32.6 & \textbf{92.1} & 48.2 & 28.9 & \textbf{87.1} & 43.4 & 51.5 & \textbf{92.7} & 66.2 & 40.3 & \textbf{91.2} & 55.3 \\
ZSC & ViL / BL & 81.0 & 76.6 & 78.7 & \textbf{75.3} & 76.7 & \textbf{76.0} & 85.9 & 79.7 & 82.7 & 81.0 & 76.6 & 78.7 & \textbf{75.3} & 76.7 & \textbf{76.0} & 89.2 & 78.0 & \textbf{83.2} & 82.8 & 77.7 & 80.1 \\
\midrule
Cluster      & - &  8.0 & 53.0 & 15.0 & 6.0 & 50.0 & 10.0 & 5.0 & 61.0 & 9.0 & 8.0 & 51.0 & 13.0 & 5.0 & 64.0 & 10.0 & 5.0 & 50.0 & 10.0 & 5.8 & 56.6 & 10.6 \\
SHOT \cite{liang+2020icml} & - & 66.3 & 51.1 & 57.7 & 59.3 & 31.0 & 40.8 & 85.8 & 31.6 & 46.2 & 73.5 & 50.6 & 59.9 & 65.3 & 28.9 & 40.1 & 84.4 & 28.2 & 42.3 & 74.6 & 33.9 & 45.6 \\
AaD \cite{yang+2022neurips} & - & 46.9 & 83.1 & 60.0 & 45.0 & 72.6 & 55.6 & 69.0 & 72.3 & 70.6 & 56.0 & 77.4 & 65.0 & 48.3 & 67.6 & 56.4 & 67.7 & 69.3 & 68.5 & 58.0 & 72.1 & 63.6 \\
UPUK \cite{wan+2024cvpr} & - &  55.9 & 85.6 & 67.6 & 45.4 & 70.2 & 55.1 & 73.9 & 83.9 & 78.6 & 56.7 & 84.1 & 67.8 & 49.3 & 74.6 & 59.4 & 69.5 & 80.0 & 74.4 & 61.6 & 79.7 & 69.2 \\
\midrule
\rowcolor{red!15}
ViLAaD++ & ViL / C-B32 & 64.6 & 71.5 & 67.9 & 59.3 & 60.7 & 60.0 & 77.1 & 70.2 & 73.5 & 69.2 & 69.6 & 69.4 & 62.2 & 63.2 & 62.7 & 81.1 & 68.2 & 74.1 & 70.5 & 66.6 & 68.4 \\
\rowcolor{red!15}
ViLAaD++ & ViL / C-L14 & 69.0 & 76.3 & 72.5 & 71.3 & 65.0 & 68.0 & 85.2 & 71.5 & 77.8 & 75.9 & 72.8 & 74.3 & 66.3 & 74.9 & 70.4 & 87.9 & 70.9 & 78.5 & 76.4 & 71.8 & 73.8 \\
\rowcolor{red!15}
ViLAaD++ & ViL / C-L14@336 & 66.6 & 78.7 & 72.2 & 68.0 & 66.5 & 67.2 & 80.4 & 74.0 & 77.1 & 75.0 & 74.7 & 74.9 & 68.4 & 73.5 & 70.9 & 87.9 & 71.6 & 79.0 & 76.3 & 72.1 & 73.9 \\
\rowcolor{red!15}
ViLAaD++ & ViL / AL & 57.1 & 82.6 & 67.6 & 42.3 & 76.0 & 54.3 & 76.0 & 69.1 & 72.4 & 61.2 & 81.8 & 70.0 & 47.4 & 75.8 & 58.4 & 76.7 & 74.8 & 75.7 & 64.8 & 74.7 & 68.5 \\
\rowcolor{red!15}
ViLAaD++ & ViL / BL & 63.3 & 78.0 & 69.9 & 61.1 & 66.0 & 63.5 & 78.6 & 72.8 & 75.6 & 69.8 & 76.3 & 72.9 & 71.7 & 66.4 & 68.9 & 81.8 & 82.5 & 82.1 & 73.8 & 72.5 & 72.9 \\
\bottomrule
\end{tabular}}
\caption{
Open-set SFDA (O-SFDA) results on Office-Home.
The table is divided into four blocks:
(1) Source model baseline,
(2) ZSC using ViL models,
(3) SFDA without auxiliary resources,
(4) ViLAaD++ (proposed).
The best results are highlighted in bold.
}
\label{tab:eval:open:oh}
\end{table*}
\section{Conclusion}
\label{sec:conc}
In this paper, we proposed ViL-enhanced AaD (ViLAaD), a natural extension of the well-adapted SFDA method, AaD \cite{yang+2022neurips}, that leverages the capabilities of ViL models.
Through extensive experiments, we demonstrated that ViLAaD outperforms both AaD and ViL-based ZSC, provided that both the source and ViL models serve as reasonable initializations for the target dataset.
To further enhance performance, we introduced ViLAaD++, which incorporates (1) an alternating optimization framework that jointly tunes ViL prompts and adapts the target model, and (2) additional loss terms designed to further refine adaptation performance.
Comprehensive evaluations on the Office-31, Office-Home, VisDA-C, and DomainNet-126 datasets show that ViLAaD++ achieves state-of-the-art results across various SFDA scenarios, including Closed-set SFDA, Partial-set SFDA, and Open-set SFDA.
\par 
As discussed in \S\ref{sec:eval:po}, despite using smaller classifiers, ViLAaD++ tends to lag behind ZSC performance with ViL models in the Open-set SFDA scenario.
In our future work, we aim to extend our approach to further improve performance in such scenarios, including Open-set SFDA and more challenging Universal SFDA \cite{schlachter+2025wacv}.





\bibliographystyle{IEEEtran}
\bibliography{bibtex/bib/IEEEexample}
\end{document}